\documentclass[journal,12 pt, onecolumn]{IEEEtran}

\IEEEoverridecommandlockouts                              

\usepackage{moreverb}

\usepackage[hyphens]{url}
\usepackage[breaklinks=true,colorlinks,bookmarksopen,bookmarksnumbered,citecolor=red,urlcolor=red]{hyperref}
\usepackage[hyphenbreaks]{breakurl}

\usepackage{graphics} 

\usepackage{amsmath}
\usepackage{cancel} 
\newcommand{\J}{\mathcal{J}}

\usepackage{my_SIunits} 
\usepackage{my_acronyms}
\usepackage{hyperref}
\usepackage{balance}
\usepackage{amsfonts}
\usepackage[]{color,graphicx}
\usepackage{subfigure} 
\usepackage{graphicx}
\usepackage{epstopdf}
\newcommand{\tflag}[1]{}
\newcommand{\tfflag}[1]{}

\usepackage{natbib}

\begin{document}

	\title{Online 3-Axis Magnetometer Hard-Iron and
		Soft-Iron Bias and Angular Velocity Sensor Bias Estimation Using Angular Velocity Sensors for
		Improved Dynamic Heading Accuracy}
	
	\author{Andrew R. Spielvogel$^{1,2}$, Abhimanyu S. Shah$^{1}$, and Louis L. Whitcomb$^{1,3}$
	\thanks{$^{1}$ Johns Hopkins University, Department of Mechanical Engineering, Baltimore, MD, USA}
	\thanks{$^{2}$ Charles Stark Draper Laboratory, Cambridge, MA, USA}
	\thanks{$^3$ Corresponding Author: {\tt llw@jhu.edu}}
        \thanks{Preprint of an article accepted for publication in \href{https://FieldRobotics.net}{Field Robotics}, \url{https://FieldRobotics.net}, Special Issue in Unmanned Marine Systems.  Submitted January 16, 2021; Revised May 28, 2021; Accepted August 2, 2021.}}



\maketitle


\begin{abstract}

This article addresses the problem of dynamic on-line estimation and
compensation of hard-iron and soft-iron biases of 3-axis magnetometers
under dynamic motion in field robotics,
utilizing only biased measurements from a 3-axis magnetometer and a
3-axis angular rate sensor.
The proposed magnetometer and angular velocity bias estimator (MAVBE) utilizes a 15-state process model encoding the nonlinear
process dynamics
for the magnetometer signal subject to angular velocity
excursions, while simultaneously estimating 9 magnetometer bias
parameters and 3 angular rate sensor bias parameters, within an
extended Kalman filter framework. 
Bias parameter local observability is numerically evaluated. \tfflag{R2-3}
The bias-compensated signals, together with 3-axis accelerometer signals,
are utilized to estimate bias compensated magnetic geodetic heading. 
Performance of the proposed MAVBE method is evaluated in comparison to the widely cited magnetometer-only TWOSTEP method in numerical simulations, laboratory experiments, and full-scale field trials of an instrumented autonomous underwater vehicle in the Chesapeake Bay, MD, USA.
For the proposed MAVBE,
$(i)$ instrument attitude is not required to estimate biases, and the results show that
$(ii)$ the biases are locally observable \tfflag{R2-3},
$(iii)$ the bias estimates converge rapidly to true bias parameters,
$(iv)$ only modest instrument excitation is required for bias estimate convergence,
and $(v)$ compensation for magnetometer hard-iron and soft-iron biases dramatically
improves dynamic heading estimation accuracy.
  
\end{abstract}


\section{Introduction}

\begin{figure}[t]
	\centering	
	\includegraphics[width=.5\textwidth]{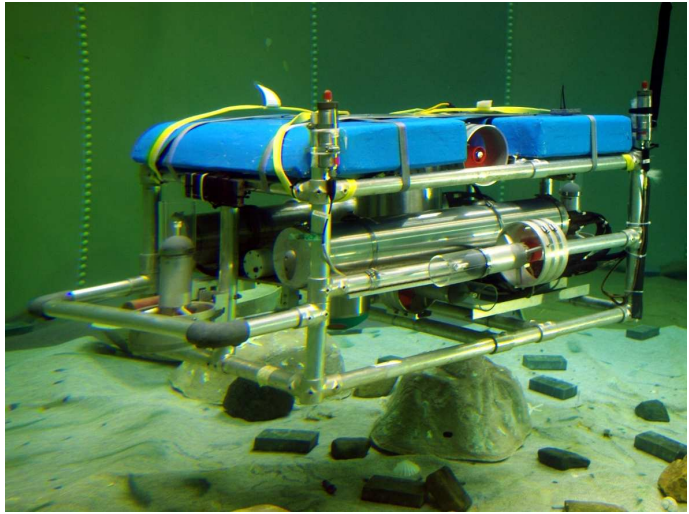}
	\caption{Full-scale experimental trials were conducted with the \acf{JHU} \acf{ROV} in a laboratory test tank (diameter of 7.5m and height of 4m). The \ac{JHU} \ac{ROV} has a full suite of navigation sensors, including several \acp{IMU}, typically found on deep submergence underwater vehicles.}
	\label{fig:tank}
\end{figure}

The dynamic instrumentation and estimation of vehicle attitude, especially geodetic heading, is
critical to accurate navigation of land, sea, and air vehicles in dynamic motion.
The utility of the ubiquitous 9-axis \acf{IMU} (with 3-axis magnetometers, 3-axis angular rate sensors, and 3-axis accelerometers) for accurate
heading  estimation is commonly vitiated by very significant hard-iron and soft-iron
magnetometer biases, as well as by  angular-rate sensor biases.
Many previously reported bias estimation approaches are complicated by
$(i)$ the need  to know the instrument's real-time attitude  (heading, pitch, and roll),
or $(ii)$ the need for the instrument to experience very large attitude
motion excursions (which may be 
infeasible for instruments mounted in many full-scale vehicles).

This article reports a novel method for dynamic on-line estimation of
hard-iron and soft-iron biases of 3-axis magnetometers under {\it
  dynamic} motion (rotation and translation) without any knowledge of
the instrument's real-time attitude.
Our approach is to formulate a nonlinear process dynamics model for
the variation in the magnetic field vector over time as the instrument
is subject to {\it a priori} unknown angular velocities.
We report a 15-state bias estimator utilizing this process model that
simultaneously estimates
$(i)$ a 3-axis dynamic process-model estimate of the true magnetic field vector,
$(ii)$ all 6 soft-iron bias magnetometer bias terms,
$(iii)$ all 3 hard-iron bias magnetometer bias terms, and
$(iv)$ all 3 angular velocity sensor bias terms.
The proposed magnetometer and angular velocity bias estimator (MAVBE) is implemented as an \acf{EKF} in which the difference between the
 estimated process model magnetometer measurement and the
actual observed magnetometer measurement provide the \ac{EKF} innovations.
The 3-axis accelerometer signal is then utilized with the estimator
signals to provide improved accuracy gyro-stabilized dynamic heading
estimation. Note that the present study does not address accelerometer bias estimation, which has been addressed in previously reported studies, e.g. \citep{batista2011,troni.TOM-2019}.   

\begin{figure*}
	\centering
	\begin{minipage}{0.49\textwidth}\subfigure[]{
			\includegraphics[width=\textwidth]{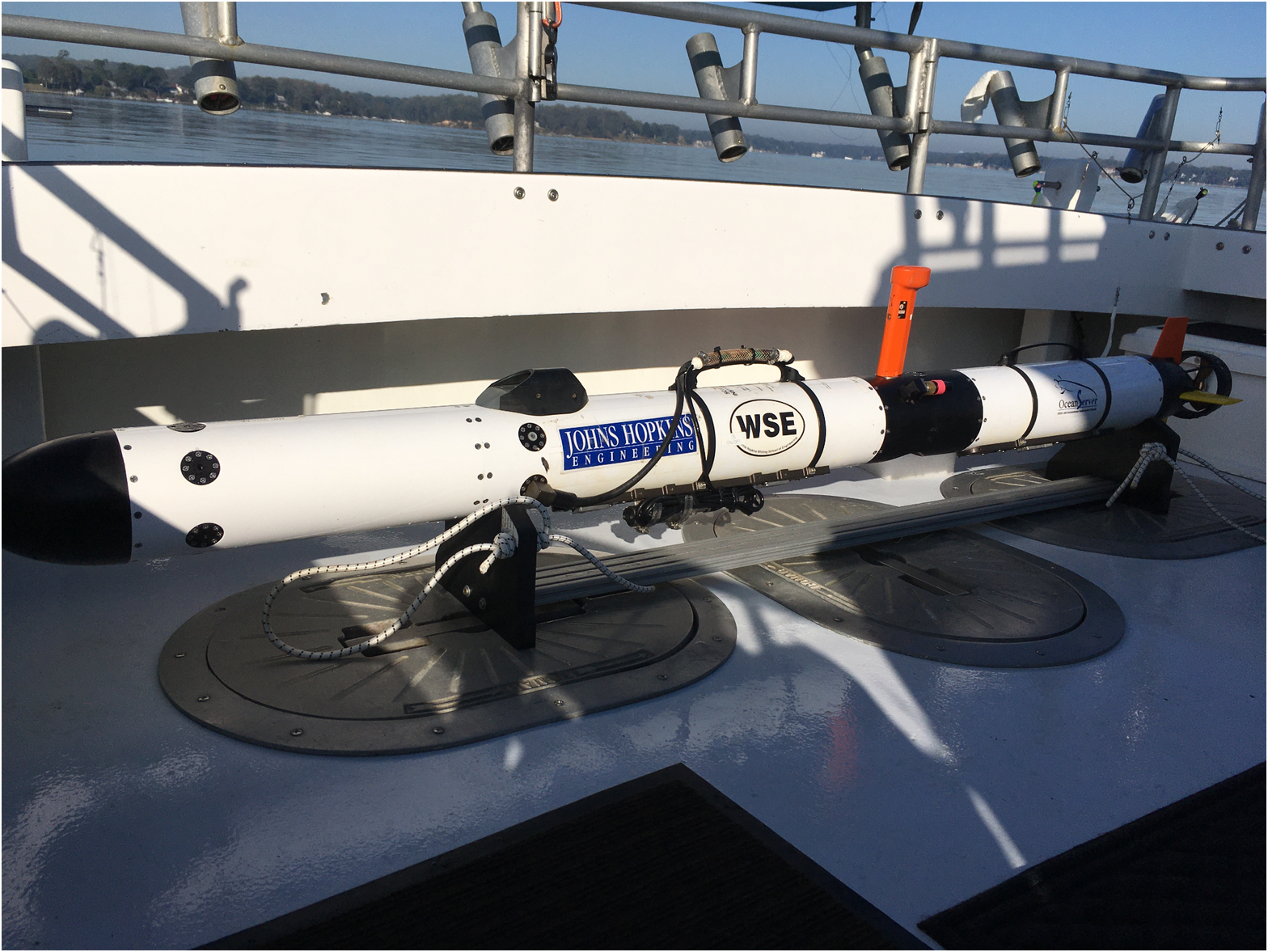}}
	\end{minipage}%
	\hfill
	\begin{minipage}{0.49\textwidth}\subfigure[]{
			\includegraphics[width=\textwidth]{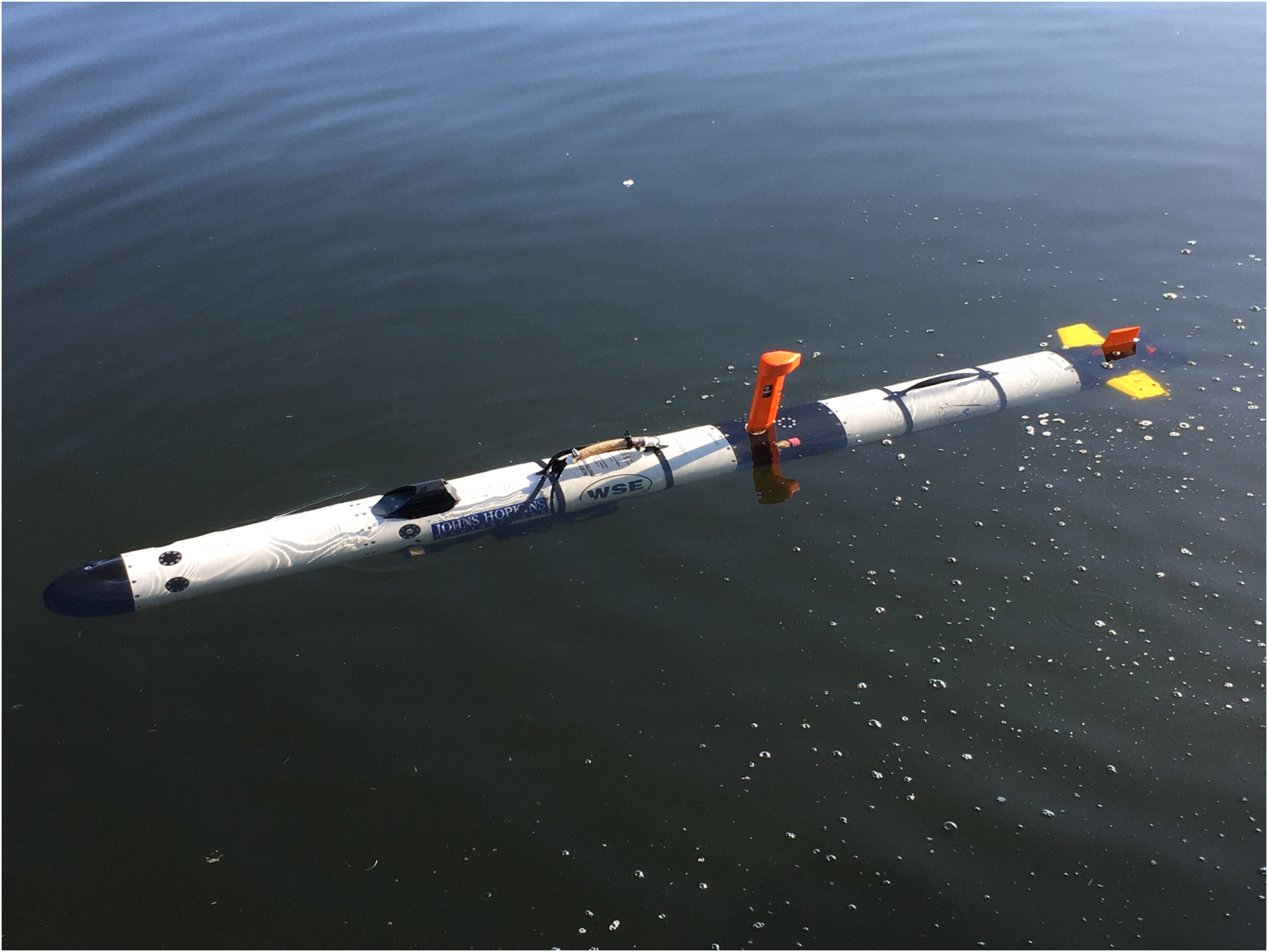}}
	\end{minipage}%
	\caption{Field trials were conducted with the \acf{JHU} Iver3 \acf{AUV} in the Chesapeake Bay, MD, USA \citep{iver3.manual}. The \ac{JHU} Iver3 has a full suite of navigation sensors, including a \acf{DVL} and several \acp{IMU} (Image credit: Paul Stankiewicz, JHU).}
	\label{fig:iver}
\end{figure*}

The present study reports a performance analysis and comparison to the widely cited magnetometer-only TWOSTEP method \citep{alonso2002complete}. The reported MAVBE is compared to the TWOSTEP method in a numerical simulation study, in full-scale laboratory experimental trials with a 9-axis \ac{IMU} on the \acf{JHU} \acf{ROV}, Figure \ref{fig:tank}, and in at-sea field experimental trials with a 9-axis \ac{IMU} on the \ac{JHU} Iver3 \acf{AUV} in the Chesapeake Bay, MD, USA, Figure \ref{fig:iver}.

Advantages of the proposed MAVBE approach include the following:  
$(i)$ knowledge of the instrument attitude is not required for sensor bias estimation,
$(ii)$ the system is shown numerically to be locally observable \tfflag{R2-3},
$(iii)$ bias estimates converge rapidly to true bias parameters,
$(iv)$ only modest instrument excitation is required for bias estimate convergence,
$(v)$ magnetometer hard-iron and soft-iron bias compensation is shown to dramatically
improve dynamic heading estimation accuracy.



\subsection{Background and Motivation}
Accurate sensing and estimation of attitude (i.e. geodetic heading, and roll and pitch referenced to the local gravitational field) are critical components of navigation systems for a wide variety of robotic vehicles.  
The need for accurate attitude estimation is
particularly acute in the case of vehicles operating in
\ac{GPS}-denied environments such as underwater.

The development of a new generation of small low-cost \acp{UV} 
has begun to enable oceanographic, environmental assessment, and national security 
missions that were previously considered impractical or infeasible \citep{clegg2003,clem2012,corke2007experiments,dunbabin2005hybrid,packard2013,steele2012,zhou2014}. These small low-cost robotic vehicles commonly employ \ac{MEMS} \acp{IMU} comprised of 3-axis \ac{MEMS} magnetometers, angular rate sensors, and accelerometers to estimate local magnetic heading, pitch, and roll, typically to within several degrees of accuracy, but require careful soft-iron and hard-iron calibration and compensation to achieve these accuracies \citep{crassidis2007survey,guo2008soft,hamel2006attitude,mahony2008,mahony2012multirotor,metni2005,metni2006,wu2015}.  Moreover, magnetic attitude sensors must be recalibrated for soft-iron and hard-iron biases whenever the vehicle's physical configuration changes significantly (i.e. sensors or other payloads added or removed), as very commonly  occurs on oceanographic marine vehicles. Studies have shown that the accuracy of magnetic heading sensors is often the principal error source in overall navigation solutions \citep{Kinsey2004DVLNAV}. In addition, \acp{EKF} rely on the assumption that noise is zero-mean Gaussian. The addition of uncompensated sensor biases introduces non-zero-mean noise that corrupts the accuracy of navigation \acp{EKF} \citep{barfoot2017state}. Thus it is essential to  estimate accurately and compensate for attitude sensor biases in order to achieve high accuracy attitude estimation.  \tfflag{R2-6} However, most previously reported approaches for hard-iron and soft-iron bias calibration require significant angular motion of the instrument in all three degrees of freedom (roll, pitch, heading) which can be impractical or infeasible on many \acp{UV} and land vehicles that are passively stable in roll and pitch. These vehicles may be unable to achieve the large roll and pitch rotations needed for traditional magnetometer calibration methods. We show that the use of angular-rate signals in the proposed MAVBE method enables the calibration to be performed with small vehicle changes in roll and pitch, thus enabling the calibration of land and underwater vehicle compasses.

\subsection{Literature Review}
 
The present study addresses the problem of  \ac{IMU} sensor bias estimation
and attitude estimation {\it utilizing only the signals from the \ac{IMU} itself}.
This study does not address the different problem of achieving \ac{IMU} sensor bias estimation/compensation and full-state and attitude estimation with
multi-sensor fusion of
\acp{IMU} {\it in combination with with additional external sensors}, which  has been addressed
in previous studies, for example for
\ac{IMU}+\ac{GPS}, e.g. \citep{Oliveria_Complementary_GPS_IMU_TCST_2011},
\ac{IMU}+computer vision, e.g. \citep{scaramuzza2014vision}, and
\ac{IMU}+Lidar, e.g. \citep{bry2012state}.
The motivation for the current study's focus on \ac{IMU} sensor bias estimation utilizing only signals from the \ac{IMU} is that
many field robots, e.g. fully submerged \acp{UV}, generally do not have access to additional external sensor signals such as from \ac{GPS}, cameras, or Lidar.


%
\tfflag{R1-1} Previously reported methods for estimating magnetometer
hard-iron biases only have been reported by Wu \citep{wu2019} and
Fedele et. al \citep{fedele2018}, but they do not address the
estimation of soft-iron magnetometer bias.  The estimation of hard-iron bias only differs from the problem solved by the present paper which
addresses both hard-iron and soft-iron bias estimation.

Many batch methods for magnetometer estimation are reported in the literature. Geophysics researchers commonly use batch
methods for magnetometer calibration
\citep{bronner2013,honsho2013}.
\tfflag{R1-1}
Zhu et. al \citep{zhu2019} formulate the calibration as a least-squares problem and
require the integration of angular rates over time to create a
rotation matrix which is used to rotate signals back to the initial
instrument frame. This integration, however, introduces random walk
errors which grow with time. 
%
In consequence, this approach can only provide coarse
magnetometer calibraion; it cannot provide fine magnetometer calibration.  
This limitation is noted by the authors, where they observe that the resulting
magnetometer calibration  
``can serve as a
coarse calibration method alone or be used as a good initial value
for other fine algorithms for more accurate result'' \citep{zhu2019}.

Alonso and Shuster \citep{alonso2002twostep} propose the TWOSTEP method for estimating magnetometer sensor bias, and an extended method \citep{alonso2002complete} for calculating magnetometer scale and orthogonality factors as well.  Vasconcelos et al. \citep{vasconselos2011} report magnetometer bias estimation as an ellipsoid fitting problem which can be solved with an iterative \ac{MLE} approach.  Many least squares methods are reported for the ellipsoid fitting problem \citep{dinale2013,ammann2015,fang2011,foster2008,ousaloo2017} and Wu et al. \citep{wu2013} frame the ellipsoid fitting problem as a particle swarm optimization (PSO). Kok et. al \citep{kok2012calibration} and Li and Li \citep{li2012new} fuse accelerometer measurements with magnetometer measurements to estimate magnetometer sensor bias, and Papafotis and Sotiriadis \citep{papafotis2019} report an algorithm for three-axis accelerometer and magnetometer calibration using a gradient descent method. All of these methods, however, require large angular rotations of the instrument
to achieve accurate bias estimation (which is infeasible for instruments mounted in many full-scale vehicles) and, moreover, they are batch estimators that are not designed  for on-line estimation of magnetometer sensor bias.

Sensor biases change over time due to changes in the configuration and payloads of the host vehicle, temperature, etc., which make it imperative to estimate sensor biases in real time. Troni and Whitcomb \citep{troni.TOM-2019} report a novel method utilizing angular velocity measurements for estimating magnetometer hard-iron sensor biases, but this approach does not address soft-iron calibration and it assumes that the
angular velocity sensor signal is already bias-compensated. Spielvogel and Whitcomb \citep{spielvogel2018iros} extend the work by Troni and Whitcomb \citep{troni.TOM-2019} to include estimation of angular-rate gyroscope and accelerometer measurement biases, but this approach again does not address soft-iron magnetometer bias calibration.

Spielvogel and Whitcomb \citep{spielvogelral2020} report an adaptive observer for magnetometer hard-iron and soft-iron biases in two-axis magnetometers. However, this observer does not provide the full calibration of 3-axis magnetometers and is only suitable for robotic vehicles which experience small roll and pitch rotations. 

Crassidis et al. \citep{crassidis2005real} report an extension to
Alonso and Shuster's TWOSTEP method, \citep{alonso2002complete,alonso2002twostep},
based on the \ac{EKF}.
Guo et al. \citep{guo2008soft} report an alternative \ac{EKF} approach for doing magnetometer sensor bias estimation.
However, these approaches require large angular rotations of the instrument for accurate magnetometer calibration, similar to the batch methods mentioned above.

Soken and Sakai \citep{soken2019} report a magnetometer calibration method using the TRIAD algorithm and an \ac{UKF}. However, this method requires knowledge of the initial attitude of the instrument and exhibits a lengthy convergence time.

Han et al. \citep{han2017} report a gyroscope-aided \ac{EKF} method for magnetic calibration. However, based upon our review of source code kindly provided by the original authors, this approach appears to integrate the angular rate forward in time which introduces random walk, thus it is not clear that it can be used continuously for a long period of time. In addition, their algorithm requires large angular rates for the hard-iron bias to converge and appears to be unable to identify the hard or soft-iron biases of high-end \ac{MEMS} \acp{IMU} like the ones used in the current paper.

\tfflag{R1-1} The study by Wu et al. \citep{wu2018} reports an error-state
 EKF for estimating 9-DOF IMU sensor biases including the hard
and soft-iron biases of a magnetometer, and sensor rotation between the
magnetometer sensor and the ``inertial sensors'' (which are the accelerometer and angular rate sensor).
This approach, an error-state EKF, differs from the full-state EKF reported in our
study.

Previous studies by the present authors on \ac{IMU} bias estimation differs from the present article in the following ways:

\begin{itemize}
	\item Two papers by Spielvogel and Whitcomb \citep{spielvogel2018rss,spielvogel_ijrr_2019} do not address magnetometer calibration and instead focus on the calibration of high-end 6-axis \acp{IMU} (\acp{IMU} with a 3-axis accelerometer and a 3-axis optical angular rate sensor) used in gyrocompass systems. 
	\item Spielvogel and Whitcomb \citep{spielvogel2018iros} extends the work by Troni and Whitcomb \citep{troni.TOM-2019} to 9-axis \acp{IMU}. However, the work estimates magnetometer hard-iron bias only, neglecting soft-iron calibration.
	\item Spielvogel and Whitcomb \citep{spielvogelral2020} report an observer for magnetometer hard-iron and soft-iron bias estimation in 2-axis magnetometers. This observer, however, is not suitable for the full 3-axis calibration of magnetometers on many robotic vehicles.
\end{itemize}  

The present article reports a novel method for real-time estimation and compensation of 3-axis magnetometer soft-iron and hard-iron and angular rate sensor biases utilizing biased angular rate sensor measurements. The proposed MAVBE algorithm requires smaller angular motion compared to previously reported magnetometer bias estimation methods, does not require any knowledge of the instrument's attitude, can be implemented on-line in real-time, exhibits rapid estimate convergence, and requires only knowledge of the local magnetic field magnitude. If the local magnetic field magnitude is not known, the proposed method can still recover the direction of the magnetic field vector which is all that is needed for accurate navigation.

\subsection{Paper Outline}

This article is organized as follows: Section \ref{sec:prelims} gives an overview of mathematical operators and definitions used in this article. Section \ref{sec:mavbe} reports the exact deterministic process model, a discrete-time stochastic approximate process model, and an \ac{EKF} formulation of the MAVBE. Section \ref{sec:eval} discuses the evaluation methodology of the MAVBE. Section \ref{sec:sim} reports a numerical simulation evaluation of the MAVBE approach for magnetometer and angular velocity sensor bias estimation, bias compensation, and heading estimation.  Section \ref{sec:tank} reports laboratory experimental evaluation of the MAVBE approach in full-scale \ac{UV} laboratory experimental trials. Section \ref{sec:field} reports field experiments of the MAVBE method in full-scale \ac{AUV} field experiments in the Chesapeake Bay, MD, USA. Section \ref{sec:conc} summarizes and concludes.

\section{Math Preliminaries}\label{sec:prelims}

\subsection{Operators}
The following operators will be used in the paper.

\emph{Skew-Symmetric Operator:} $\mathcal{J}$ is a mapping $\mathbb{R}^3 \rightarrow so(3)$, such that $\forall \ x \in \mathbb{R}^3$,
\begin{equation}
\mathcal{J}(x) = 
\begin{bmatrix}
0 & -x_3 & x_2 \\
x_3 & 0 & -x_1 \\
-x_2 & x_1 & 0
\end{bmatrix}\ , 
\end{equation} 
where $x = [x_1\ x_2\ x_3]^T$. 

\emph{Jacobian Operator:} $\mathbf{D}$ is a mapping $\mathbb{R}^m \rightarrow \mathbb{R}^{m \times n}$ , such that for $x \in \mathbb{R}^n$ and $f:\mathbb{R}^n\rightarrow \mathbb{R}^m$, $\mathbf{D}_x[f(x)]_{\mu}$ gives the $m\times n$ Jacobian of $f(x)$ with respect to $x$ evaluated at $\mu$.

\emph{Stack Operator: } ($\cdot)^s$ is a mapping $\mathbb{R}^{m \times n} \rightarrow \mathbb{R}^{mn}$.
For a matrix  $A \in \mathbb{R}^{m \times n}$, the stack operator \citep{schacke2013} is defined as
\begin{equation}
A^s = [a_{11}\ \hdots\ a_{m1}\ a_{12}\ \hdots\ a_{m2}\ \hdots\ a_{1n}\ \hdots\ a_{mn}]^T.
\end{equation}

\subsection{Kronecker Product} 
The Kronecker product \citep{LOAN200085,schacke2013} of matrix $A \in \mathbb{R}^{p \times q}$ and $B \in \mathbb{R}^{r \times s}$, denoted $A \otimes B \in \mathbb{R}^{pr \times qs}$, is defined as \tfflag{R4-5}
\begin{equation}
A \otimes B = 
\begin{bmatrix}
a_{11}B & \hdots & a_{1q}B \\
\vdots & & \vdots \\
a_{p1}B & \hdots & a_{pq}B
\end{bmatrix}.
\end{equation} 

\subsection{Definitions}

\emph{Persistent Excitation (PE) \citep{narendra2012stable,sastryadaptive}:} A matrix function $\mathcal{W}:\mathbb{R}^+\rightarrow\mathbb{R}^{m\times n}$ is persistently exciting (PE) if there exist $T,\alpha_1,\alpha_2>0$ such that $\forall t\geq0$:
	\begin{align}
	\alpha_1I_m\geq\int_{t}^{t+T}\mathcal{W}(\tau)\mathcal{W}^T(\tau)\,d\tau\geq\alpha_2I_m
	\end{align}
	where $I_m\in\mathbb{R}^{m\times m}$ is the identity matrix.

\section{Magnetometer and Angular Velocity Bias Estimator (MAVBE) Formulation}\label{sec:mavbe}
This section reports the derivation of a novel online magnetometer and angular velocity bias estimator (MAVBE) for simultaneous hard-iron and soft-iron magnetometer and angular-rate sensor bias estimation and compensation. The  biases are assumed to be very slowly time varying, and hence we model them as constant terms and update the estimates continuously.

\subsection{Exact System Process Model}
Magnetometers (including those employed in \acp{IMU}) are subject to two primary sensor calibration errors: hard-iron and soft-iron biases. Hard-iron biases are constant magnetometer sensor bias terms due to the permanent magnetic signature of the instrument and the vehicle. Soft-iron biases are non-constant magnetometer sensor bias terms due to the magnetic permeability of the instrument and the vehicle, and will vary with vehicle heading and attitude.
For most \ac{IMU} magnetometers, hard-iron biases dominate soft-iron biases.

The most commonly utilized models for 3-axis magnetometer hard-iron and soft-iron bias, and for
3-axis angular velocity sensor bias are
\begin{align}    
m_m(t) &= T m_t(t) + m_b  \label{eq:m_model}\\
w_m(t) &= w_t(t) + w_b  \label{eq:w_model},
\end{align}
where $m_m(t)\in\mathbb{R}^{3}$ and $m_t(t)\in\mathbb{R}^{3}$ are the noise-free measured and true magnetometer values, respectively, in the instrument frame, $T\in\mathbb{R}^{3\times 3}$ is a \ac{PDS} matrix due to soft-iron effects, $m_b\in\mathbb{R}^{3}$ is the sensor bias due to hard-iron effects, $w_m(t)\in\mathbb{R}^{3}$ and $w_t(t)\in\mathbb{R}^{3}$ are the noise-free measured and true angular velocity signals, respectively, in the instrument frame, and $w_b\in\mathbb{R}^{3}$ is the angular velocity bias.

The \ac{PDS} matrix $T$ is parameterized as
\begin{align}
T&=\left[\begin{array}{ccc}
a&b&c\\b&d&e\\c&e&f
\end{array}\right]
\end{align}
where we define $t_p\in\mathbb{R}^6$ as the vector of the 6 unique elements of $T$ such that 
\begin{align}
t_p&=\left[\begin{array}{cccccc}
a&b&c&d&e&f
\end{array}\right]^T. \label{eq:tp}
\end{align}

The true magnetometer value in the \ac{NED} frame, $^wm_t\in\mathbb{R}^{3}$, is constant and is related to the true magnetometer value in the instrument frame by 
\begin{equation}\label{eq:model2}
^wm_t = R(t)\; m_t(t),
\end{equation}
where $R(t) \in SO(3)$ is the time varying rotation of the instrument frame with respect to the \ac{NED} frame. Taking the time derivative of (\ref{eq:model2}) yields
\begin{equation}\label{eq:model4}
0 = \dot{R}(t)\ m_t(t) + R(t)\dot{m}_t(t).
\end{equation}
Rearranging (\ref{eq:model4}) yields
\begin{equation}\label{eq:model5}
\dot{m}_t(t) = - \mathcal{J}(w_t(t))m_t(t),
\end{equation}
and substituting (\ref{eq:w_model}) yields
\begin{equation}\label{eq:model6}
\dot{m}_t(t) = -\mathcal{J}(w_m(t) - w_b) m_t(t).
\end{equation}

Since the bias terms are assumed to be at most very slowly time varying, they are modeled as constants. Finally, the full state process model can be written in state space form as
\begin{align}
\underbrace{
	\left[
	\begin{array}{c}
	\dot{m}_t(t) \\
	\dot{m}_b    \\
	\dot{t}_p    \\
	\dot{w}_b  
	\end{array}
	\right ]}_{}
&=
\underbrace{
	\left[
	\begin{array}{c}
	-\mathcal{J}(w_m(t) - w_b) m_t(t)   \\
	\mathbb{O}_{3 \times 1}  \\
	\mathbb{O}_{6 \times 1} \\
	\mathbb{O}_{3 \times 1} 
	\end{array}
	\right ]}_{}\\
\dot{\Phi}(t)\,\,\,\,\,\,\,&= \,\,\,\,\,\,\,\,\,\,\,\,\,\,\,\,\, f\left(\Phi (t)\right)\label{eq:f(x)},
\end{align}
with the measurement model
\begin{align}
\underbrace{\left[\begin{array}{c}m_m(t)\\ \|m_t(t)\|^2\end{array}\right]}_{}&=\underbrace{\left[\begin{array}{c}
	T\ m_t(t) + m_b\\
	m_t^T(t)m_t(t)
	\end{array}\right]}_{} \label{eq:meas_model}\\
z(t)\,\,\,\,\,\,\,\,\,\,\,&=\,\,\,\,\,\,\,\,\,\,h(\Phi(t))
\end{align}
where $\mathbb{O}_{m\times n}$ is a zero matrix of dimension $m\times n$,  $\Phi(t)\in\mathbb{R}^{15}$ is the state vector, and $z(t)\in\mathbb{R}^4$ is the measurement. The $w_b$ bias term is included in the state vector because we have a process model for it (bias is assumed to be constant), however the time-varying $w_m(t)$ term was excluded from the state since its process model is unknown. It is instead considered to be an exogenous input signal. Modeling $w_m(t)$ as a exogenous input is a common approach used in magnetometer calibration (e.g. \citep{troni.TOM-2019,han2017,soken2019}. Note that unlike in the work by Troni and Whitcomb \citep{troni.TOM-2019}, herein we do not assume the angular velocity signal to be already bias-compensated. 

As seen in (\ref{eq:meas_model}), the measurement model requires knowledge of the magnitude of the local magnetic field vector which is available for most field vehicles via the World Magnetic Model (WMM) \citep{WMM} or the International Geomagnetic Reference Field (IGRF) model \citep{IGRF}. Note that if the local magnetic field vector magnitude is unknown, by setting $\|m_t(t)\|^2$ to a non-zero positive constant (e.g. Setting $\|m_t(t)\|^2=1$ will normalize the estimated corrected magnetic field vector to 1.), the direction of the corrected magnetic field vector can still recovered, allowing accurate magnetic heading estimates.

\subsection{MAVBE Process Model Linearization} \label{sec:linearization}
The MAVBE \ac{EKF} was implemented using a linear process model. The nonlinear dynamics given in (\ref{eq:f(x)}) were linearized using an approach that follows closely from that used by Webster \citep{webster.TH-2010}. The linearization of $f(\Phi)$ about an arbitrary operating point $\mu$ is given by 
\begin{equation}
f(\Phi(t)) = f(\mu) + \mathbf{D}_{\Phi(t)}[\ f(\Phi(t))\ ]_{\mu} (\Phi(t) - \mu)\ + H.O.T.
\end{equation}
Neglecting the higher order terms ($H.O.T$), and rearranging yields
\begin{equation}\label{eq:linear}
f(\Phi(t)) =  \underbrace{\mathbf{D}_{\Phi}[\ f(\Phi(t))\ ]_{\mu}}_{A(\Phi(t))} \Phi(t) + \underbrace{f(\mu) - \mathbf{D}_{\Phi}[\ f(\Phi(t))\ ]_{\mu} \mu}_{u(\Phi(t))}
\end{equation}
where $u(\Phi(t))\in\mathbb{R}^{15}$ is referred to as the pseudo-control input and $A(\Phi(t))\in\mathbb{R}^{15\times 15}$ is the Jacobian of $f(\Phi(t))$ with respect to $\Phi(t)$ evaluated at $\mu$. The noise-free linearized process model is given by 
\begin{align}
\dot{\Phi} &= A(\Phi(t))\Phi(t) + u(\Phi(t))\label{eq:state}\\
z(t)&=h\left(\Phi(t)\right) \label{eq:obs}
\end{align}
where  
\begin{align}
A(\Phi(t)) &=	\mathbf{D}_\Phi[\ f(\Phi(t))\ ]_{\mu}  \\
&= 
\left[
\begin{array}{ccc}
-\J(w_m(t) - w_b)&\mathbb{O}_{3 \times 9}&-\mathcal{J}(m_t(t))\\
\mathbb{O}_{12 \times 3}&\mathbb{O}_{12 \times 9}&\mathbb{O}_{12 \times 3}
\end{array} \right]_{\mu}.
\end{align}

\subsection{MAVBE Observability}\label{sec:observability} \tfflag{R2-3}
The linearized system (\ref{eq:state})-(\ref{eq:obs}) is locally observable on $[t_0,t_f]$ if and only if the observability Gramian
\begin{align}
M(t_0,t_f)&=\int_{t_0}^{t_f}\mathcal{H}^T(t,t_0)C^T(t)C(t)\mathcal{H}(t,t_0)\,dt
\end{align}
is full rank where $\mathcal{H}(t,t_0)$ is the state transition matrix \citep{rugh} and $C(t)$ is the measurement Jacobian
\begin{align}
C(t) &=	\mathbf{D}_{\Phi(t)}[\ h(\Phi(t))\ ]_{\Phi(t)}.
\end{align} 
\tfflag{R2-2} \tfflag{R2-3} Through numerical studies we found that a \ac{PE} $w_m(t)$ signal results in a full-rank observability Gramian, and thus local observability of the state vector. Although it is unclear how to show this result ananlytically, we were able to check numerically, that a variety of \ac{PE} $w_m(t)$ signals result in a full-rank observability Gramian, and thus local observability of the state vector.

\subsection{MAVBE Process Model Discrete-Time Stochastic Approximation}
Using the approach followed by Webster \citep{webster.TH-2010}, the noise free continuous time system (\ref{eq:state})-(\ref{eq:obs}) can be approximated as the discrete-time stochastic system \tfflag{R4-2}
\begin{align}\label{eq:discretized}
\Phi_{k} &= \bar{A}_{k-1}  \Phi_{k-1} + \bar{B}_{k-1} u_{k-1} + w_{k-1} \\
z_k &= h\left(\Phi_k\right) + v_k,
\end{align}
where $\Phi_{k}$ is $\Phi(k)$ at time $t=k$, $u_k$ is $u(\Phi(k))$, $w_k\sim \mathcal{N}(0,Q)$ is the independent zero-mean Gaussian process noise, and $v_k\sim \mathcal{N}(0,R)$ is the independent zero-mean Gaussian measurement noise.
\tfflag{R2-4} Note that we have an exact deterministic noise-free continuous-time plant model,
we do not have a continuous-time noise model for plant process noise. Here we
construct a discrete-time approximation of the noise-free continuous-time plant model,
and adopt the assumption of additive zero-mean Gaussian discrete-time process noise. The
process noise for the discrete system was chosen to be a diagonal matrix for simplicity and
ease of tuning.

$\bar{A}_k$ is related to $A(\Phi(k))$ by
\begin{equation}
\bar{A}_k = e^{A(\Phi_k)\tau}
\end{equation}
where $\tau$ is the discretization time step. Similarly, $\bar{B}_k$ can be computed as\\
\begin{equation}
\begin{split}
\bar{B}_k & = \int_{0}^{\tau} e^{A(\Phi_k)(\tau-s)}\;ds \\
& = e^{A(\Phi_k)\tau}  \int_{0}^{\tau}  e^{-A(\Phi_k)s}ds \\
& = \bar{A}_k \int_{0}^{\tau}  e^{-A(\Phi_k)s}ds.
\end{split}
\end{equation}

\subsection{MAVBE Extended Kalman Filter For Magnetometer Bias Estimation}
The process model used to predict the state is
\begin{equation}\label{eq:ekf1}
\Phi'_{k} = \bar{A}_{k-1} \Phi_{k-1} + \bar{B}_{k-1} u_{k-1},
\end{equation}
and the predicted covariance matrix is
\begin{equation}\label{eq:ekf2}
\Sigma'_{k} = \bar{A}_{k-1} \Sigma_{k-1} \bar{A}_{k-1}^{T} + Q,
\end{equation}
where Q is the process noise covariance matrix (constant). The Kalman gain is given by
\begin{equation}\label{eq:ekf3}
K_k = \Sigma'_k \bar{C_k}^T (\bar{C_k} \Sigma'_k \bar{C_k}^T + R)^{-1},
\end{equation}
where $R$ is the measurement covariance matrix (constant), and \tfflag{R4-4}
\begin{align}
\bar{C_k} &=	\mathbf{D}_\Phi[\ h(\Phi_k)\ ]_{\Phi'_k}  \\
&= 
\left[
\begin{array}{cccc}
	T&\mathbb{I}_{3}&\left(m_t^T(k)\otimes\mathbb{I}_{3}\right)D_{t_p}[T^s]&\mathbb{O}_{3\times 3}\\
	2m_t^T(k)&\mathbb{O}_{1 \times 3}&\mathbb{O}_{1 \times 6}&\mathbb{O}_{1\times 3}
\end{array} \right]_{\Phi'_k}
\end{align}
where $\mathbb{I}_{n}$ is the identity matrix of dimension $n\times n$.

The updated state estimate is \tfflag{R4-3}
\begin{equation}\label{eq:ekf4}
\Phi_k = \Phi'_k + K_k(z_k - \bar{C_k}\Phi'_k).
\end{equation}

The updated covariance estimate is
\begin{equation}\label{eq:ekf5}
\Sigma_k = (\mathbb{I}_{15} - K_k \bar{C}_k) \Sigma'_k.
\end{equation}

\section{MAVBE Performance Evaluation}\label{sec:eval}

The remainder of this paper is concerned with evaluating the
proposed  MAVBE presented in Section \ref{sec:mavbe}
in comparison to previously reported
approaches to magnetometer bias estimation.
We report a comparative performance analysis of the proposed MAVBE to
the widely cited batch-processing TWOSTEP method reported by Alonso and Shuster \citep{alonso2002complete}.
We note that Dinale \citep{dinale2013} provides an excellent overview of the TWOSTEP method. In particular, his Appendix C.1 provides a reference Matlab implementation of the TWOSTEP method which proved to be very helpful to the current authors when implementing the TWOSTEP comparison.

We evaluated the performance of these approaches in the following three ways:

\begin{enumerate}
	
	\item First, in Section \ref{sec:sim}, we report the evaluation of these
	approaches in numerical simulations in which both the the true simulated bias
	values and the true simulated heading values are known exactly.
	
	\item Second, in Section \ref{sec:tank}, we report the evaluation of these
	approaches in full-scale laboratory experimental trials with a
	laboratory testbed ROV in which the true sensor biases are unknown, and thus the accuracy of the estimated biases cannot be measured directly. Instead, the accuracy of the heading estimate is used as an error metric for sensor bias estimation. The heading estimate was computed using accelerometer and magnetometer signals from the MicroStrain 3DM-GX5-25 9-Axis IMU (LORD Sensing-MicroStrain, Williston, Vermont, USA) \citep{microstraingx3datasheet}, together with the sensor bias estimates. In the laboratory experiments, the estimated heading was compared to the ground truth heading from a high-end \ac{INS}, the iXBlue PHINS III (iXBlue SAS, Cedex, France), with heading accuracy of $0.05^\circ/\cos$(latitude) \citep{phins.specs,ixsea.phins3.manual}.
	
	\item Third, in Section \ref{sec:field}, we report the evaluation of these
	approaches in full-scale sea-trials with the \ac{JHU} Iver3 \citep{iver3.manual}, in which
	neither the true bias values nor the true heading is known. 
	In the field experimental trials, vehicle XY navigation error is used as a proxy for the accuracy of the magnetometer calibration. The vehicle track is recalculated using Doppler dead reckoning for each magnetometer calibration method and compared to the \ac{GPS} track of the vehicle. In addition to the TWOSTEP method, the proposed MAVBE is compared to L3 OceanServer's commercial solution for a calibrated magnetic compass using the OceanServer OS5000 magnetic compass \citep{os5000}.
	
\end{enumerate}

\subsection{Attitude Calculation}
The vehicle coordinates are defined such that the x-axes is pointed forward on the vehicle, the y-axes is pointing starboard, and the z-axes down. Using this coordinate frame, the instantaneous estimated roll $\phi$ and pitch $\theta$ angles of the vehicle can be computed by \citep{tronidis}
\begin{align}
\hat{\phi}&=atan2\left(-a_y,-a_z\right) ,\\
\hat{\theta} &= atan2\left(a_x,\sqrt{a^2_y+a^2_z}\right),
\end{align}
where $a_x,a_y,a_z$ are the $x,y,z$ components, respectively, of the accelerometer signal.
The calibrated magnetic field vector, $m_t(t)$, in the vehicle frame is transformed to the local level frame by the relation ${}^lm={}^l_vR(t){}m_t(t)$ where ${}^l_vR(t)\in SO(3)$ is the rotation matrix using the roll and pitch estimates. The instantaneous estimated heading can then be computed as \citep{tronidis}
\begin{align}
\hat{\gamma}&=atan2\left(-{}^lm_y,{}^lm_x\right) - \gamma_0 ,
\end{align}
where $\gamma_0$ is the known local magnetic variation and where ${}^lm_x,{}^lm_y,{}^lm_z$ are the $x,y,z$ components, respectively, of the ${}^lm$ signal.

\begin{figure*}[t]
	\centering
	\begin{minipage}{0.5\textwidth}\subfigure[{\bf Sim1}: True magnetic field vector.]{
			\includegraphics[width=\textwidth]{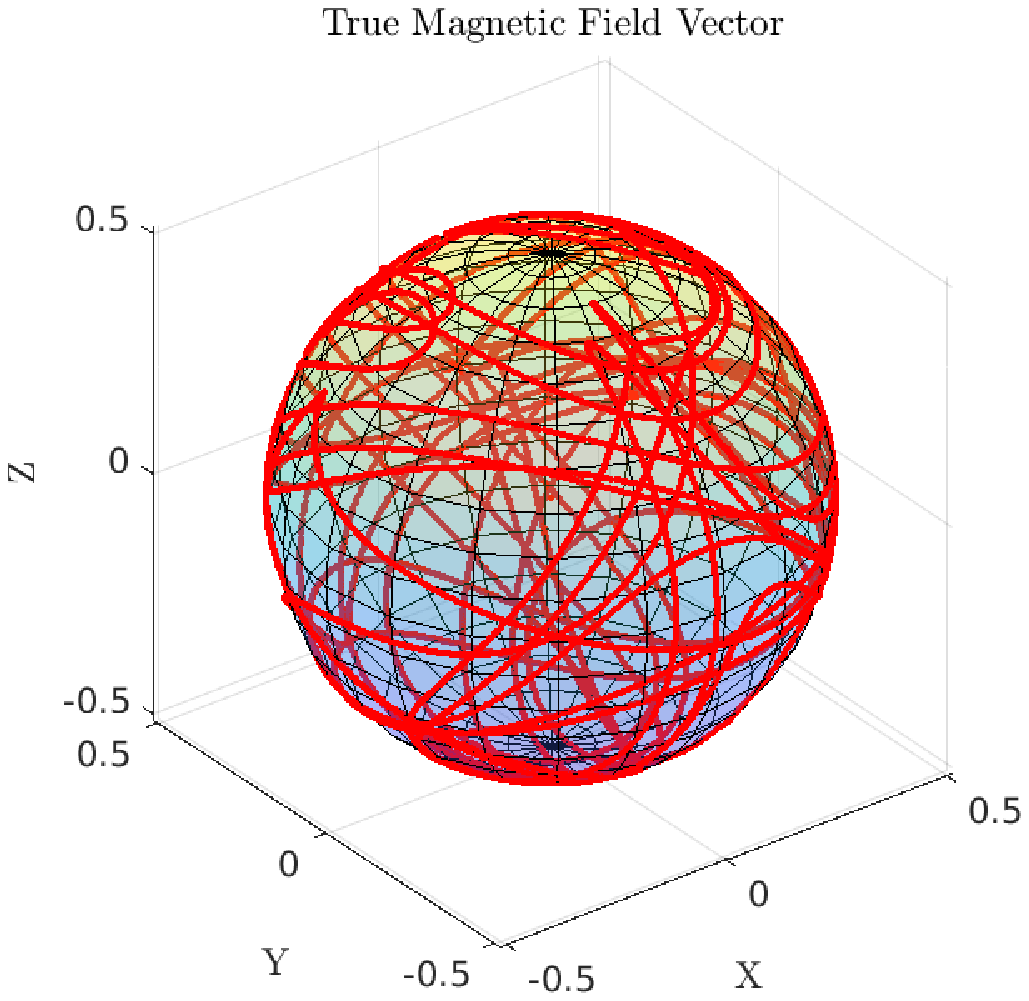}}
	\end{minipage}%
	\hfill
	\begin{minipage}{0.5\textwidth}\subfigure[{\bf Sim2}: True magnetic field vector.]{
			\includegraphics[width=\textwidth]{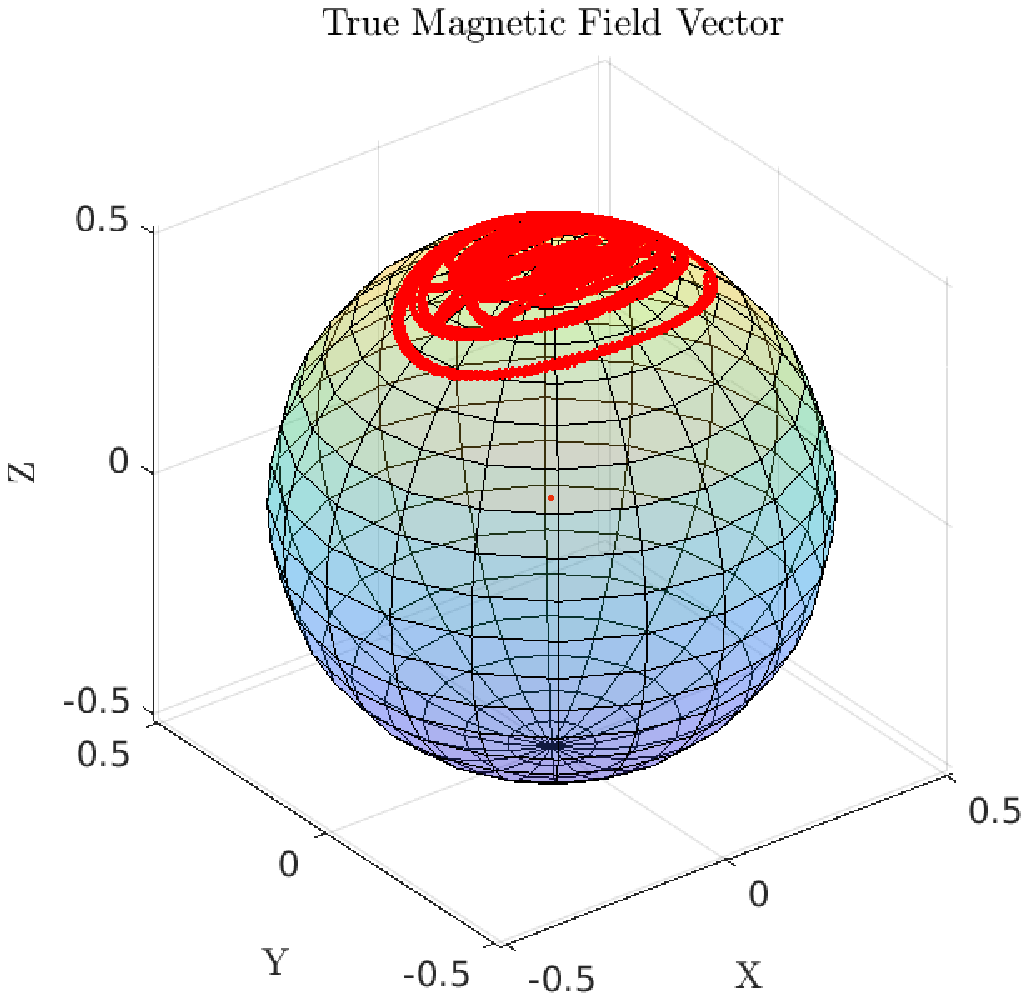}}
	\end{minipage}%
	\caption{{\bf Simulation:} The true magnetic field vectors in the instrument frame during the simulations.}
	\label{fig:mag} 
\end{figure*}
\begin{table*}[h]
	\centering
	\caption{Simulation Setup Parameters.}	
	\footnotesize
	\begin{tabular}{|c |  l|}
		\hline
		& \bf Sensor Noise \\
		\hline
		$\sigma_m$ & [2  2  2]$^T\cdot 10^{-4}$ G \\
		$\sigma_w$ & [2.4  2.4  2.4]$^T\cdot 10^{-4}$ rad/s \\
		\hline
		& \bf Sensor Bias \\
		\hline
		$m_b$ & [0.6  -0.7  -1]$^T\cdot 10^{-1}$ G \\
		$w_b$ & [-2  3  -1]$^T\cdot 10^{-3}$ rad/s \\
		$t_p$ & [1.1  0.1  0.03  0.95  0.01  1.2]$^T$ \\
		\hline
		& \bf Sensor Bias Estimate Initial Values \\
		\hline
		$\hat{m}_b(t_0)$ & [0  0  0]$^T$ G\\
		$\hat{t}_p(t_0)$ & [1  0  0  1  0  1]$^T$\\
		$\hat{w}_b(t_0)$ & [0  0  0]$^T$ rad/s\\
		\hline
	\end{tabular}
	\label{table:sim_setup}
\end{table*}
 
\subsection{Doppler Dead Reckoning Navigation}

\acfp{DVL} are commonly used on \acp{UV} to measure vehicle three-axis velocity.  When a \ac{DVL} has bottom-lock, the instrument provides accurate measurements of the \ac{UV}'s three-axis velocity with respect to the fixed sea floor. 

Using the roll, pitch, heading attitude of the vehicle, these velocity measurements can be transformed into world frame by \citep{tronidis}
\begin{align}
{}^wv(t)&={}^w_vR(t)\,{}^v_iR\,{}^iv(t)
\end{align}
where ${}^v_iR$ is the constant rotation matrix from the instrument coordinate frame to the vehicle coordinate frame, ${}^w_vR(t)$ is the time-varying rotation matrix from the vehicle coordinate frame to the inertial world coordinate frame (the rotation matrix corresponding to the roll, pitch, heading of the vehicle), ${}^iv(t)$ is the vehicle's velocity in the \ac{DVL} instrument's coordinate frame, and ${}^wv(t)$ is the world frame vehicle velocity. ${}^wv(t)$ can then be integrated to provide the dead reckoning position estimate \citep{whitcomb1999} 
\begin{align}
{}^wp(t)&={}^wp(t_0) + \int_{t_0}^{t}{}^wv(\tau)\, d\tau.
\end{align}

\section{MAVBE Simulation Evaluation}\label{sec:sim}

\subsection{Simulation Setup}
The proposed bias estimation and compensation algorithm was evaluated in a numerical simulation using Matlab. Simulated \ac{IMU} sensor measurements were generated by simulating sinusoidal vehicle motions. The simulated data was generated at $20$ hz. Noise was added to this data with characteristics observed from experimental bench tests of a MicroStrain 3DM-GX5-25 \citep{microstraingx3datasheet}. The measured standard deviations of the simulated magnetometer and angular-rate sensors sampled at $20$ hz and the simulated ``true'' sensor biases used during the simulated data generation, which are realistic sensor bias values for a MicroStrain 3DM-GX5-25 as observed from bench tests, are given in Table \ref{table:sim_setup}.

The simulated instrument experienced smooth sinusoidal rotations. In {\bf Sim1}, the instrument experiences changes in roll, pitch, and heading of $\pm180^\circ$. However, in {\bf Sim2} the instrument experiences smaller angular rotations with roll and pitch magnitudes of $<50^\circ$ and heading of $\pm 180^\circ$. Figure \ref{fig:mag} shows the true magnetic field in the instrument frame. 

The MAVBE was executed at $10$ hz, the initial conditions for the sensor biases estimates are given in Table \ref{table:sim_setup}, and the measurement covariance matrix, $R$, was populated with the square of the measured standard deviations, along the diagonal entries such that
\begin{align}
R &= diag([4\ 4\ 4])\cdot 10^{-8},  \label{eq:R}
\end{align} and a process covariance matrix, $Q$, that works well is 
\begin{align}
Q &= diag([1\ 1\ 1\ 1\ 1\ 1\ 1\ 1\ 1\ 1\ 1\ 1\ 0.01\ 0.01\ 0.01])\cdot 10^{-10}. \label{eq:Q}
\end{align}
Larger $Q$ values resulted in quicker convergence, at the expense of more oscillatory final steady states, while lower $Q$ values resulted in slower convergence, but a smoother final steady state. We selected $Q$ empirically to provide a balance between convergence time and a smooth final steady state. In the future, two different process covariance matrices could be chosen for coarse and fine alignment in order to achieve fast convergence and a less oscillatory final steady state. Also, after an initial calibration, previous estimates for the sensor biases can be used as initial conditions to the MAVBE in order to greatly decrease convergence time of the proposed estimator.

\begin{figure*}[t!]
	\centering
	\includegraphics[width=.75\textwidth]{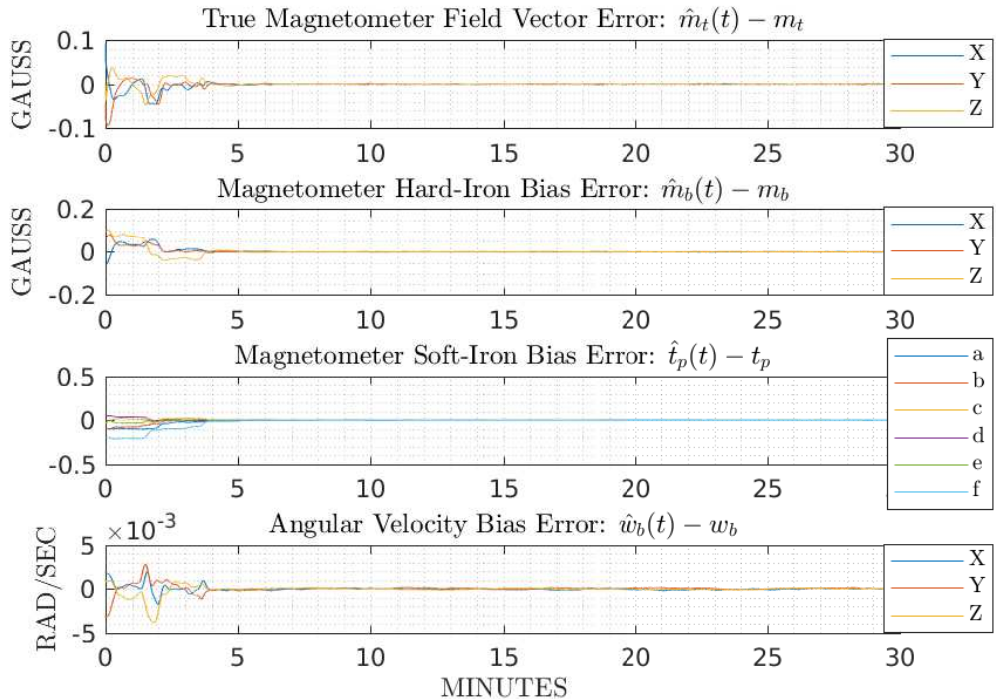}
	\caption{{\bf Simulation 1 Results.} MAVBE magnetometer and angular rate sensor bias estimate errors converge to zero, i.e the estimated biases converge to their known true values.}
	\label{fig:sim1_bias}
\end{figure*}


\begin{figure*}[t!]
	\centering
	\includegraphics[width=.75\textwidth]{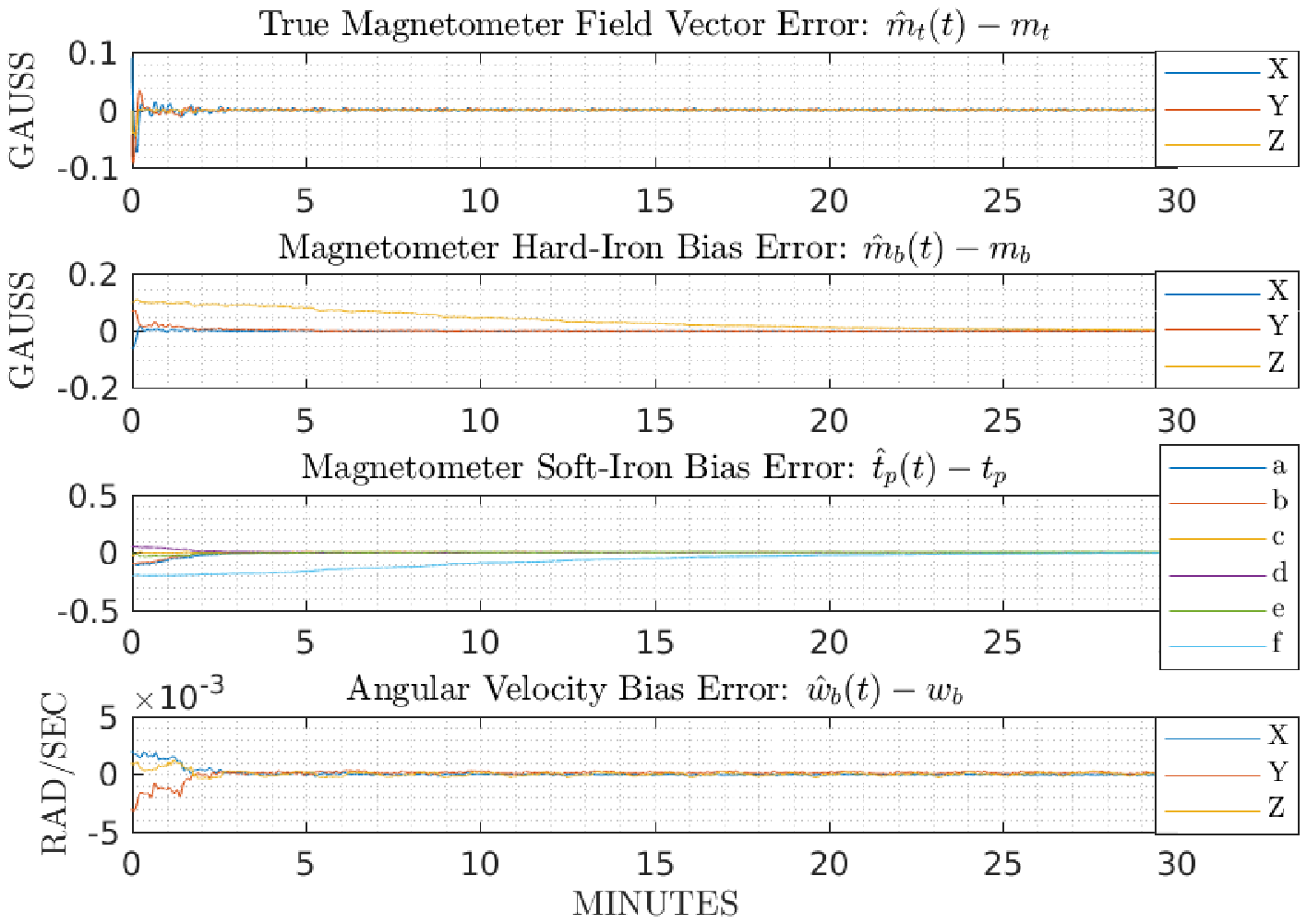}
	\caption{{\bf Simulation 2 Results.} MAVBE magnetometer and angular rate sensor bias estimate errors converge to zero, i.e the estimated biases converge to their known true values.}
	\label{fig:sim2_bias_tp}
\end{figure*}

\begin{table*}[t]
	\centering
	\caption{Estimated bias values for the two numerical simulations.}
	\footnotesize
	\begin{tabular}{|l | c c c|}
		\hline
		& $m_b$ (G)& $t_p$ & $w_b$ ($^\circ$/s) \\
		\hline
		\bf True & \bf [0.060 -0.070 -0.100] & \bf [1.100 0.100 0.030 0.950 0.010 1.200] &\bf [-0.002 0.003 -0.001]\\
		\hline 
		\bf MAVBE Sim1 & [0.059 -0.070 -0.100] & [1.099 0.100 0.030 0.949 0.010 1.200] & [-0.002 0.003 -0.001]\\
		\bf TWOSTEP Sim1 & [0.064 -0.080 -0.084] & [1.100 0.100 0.030 0.950 0.010 1.200] & N/A \\
		\hline
		\bf MAVBE Sim2 & [0.060 -0.070 -0.096] & [1.100 0.100 0.030 0.950 0.010 1.192] & [-0.002 0.003 -0.001]\\
		\bf TWOSTEP Sim2 & [0.062 -0.075 -0.123] & [1.148 0.104 0.031 0.992 0.009 1.317] & N/A \\
		\hline
	\end{tabular}
	\label{table:biases_sim}
\end{table*}

\begin{table*}[t!]
	\centering
	\caption{Comparison of heading \acp{RMSE} between the calibration methods during the numerical simulations.}	\footnotesize
	\begin{tabular}{|c | c c|}
		\hline
		\bf Calibration Method & \bf Sim1 & \bf Sim2 \\
		\hline
		\bf Uncalibrated &  $23.35^\circ$ & $19.21^\circ$ \\
		\bf MAVBE & $0.54^\circ$ & $0.58^\circ$  \\
		\bf TWOSTEP & $0.61^\circ$ & $1.42^\circ$ \\
		\hline
	\end{tabular}
	\label{table:rms_sim}
\end{table*}

\subsection{Simulation Results}
Figures \ref{fig:sim1_bias} and \ref{fig:sim2_bias_tp} show the
evolution of the error of the estimated sensor biases for the two simulation.
The simulation results show that the MAVBE can accurately estimate the
true values of $m_b$, $t_p$, and $w_b$, \tfflag{R1-2},
and that the parameter estimates converge to the true parameter values within 5-10 minutes.
The parameter convergence of the {\bf  Sim1} simulation,
shown in Figure \ref{fig:sim1_bias},
with very large $\pm180^\circ$ excursions in heading, pitch, and roll,
is about twice as fast as the
parameter convergence of the {\bf  Sim2} simulation,
shown in Figure \ref{fig:sim2_bias_tp}, 
with only $<50^\circ$ excursions in pitch, and roll,
Since, in simulation, the true values of the
biases are known, it is easy to verify that the biases estimates
converge to the true bias values, and not to arbitrary incorrect
values, which would generally be the case if $w_m(t)$ was not a
\ac{PE} signal.


Table \ref{table:biases_sim} report the biases estimated by the MAVBE and TWOSTEP methods during {\bf Sim1} and {\bf Sim2}.  In addition, Table \ref{table:rms_sim} lists the heading errors corresponding to using the biases from Table \ref{table:biases_sim} for magnetometer calibration.

In {\bf Sim1}, the resulting heading \ac{RMSE} of the calibrated magnetometer from the MAVBE and TWOSTEP methods was similar at $0.54^\circ$ and $0.61^\circ$ \ac{RMSE} respectively. In {\bf Sim1}, the large angular rotations of the instrument provide rich magnetometer measurements, allowing both of the methods to find the true sensor biases. However, in {\bf Sim2} the resulting heading \ac{RMSE} of the calibrated magnetometer from the MAVBE and TWOSTEP methods differ.  The MAVBE method performs better, with a heading \ac{RMSE} of $0.58^\circ$, than the TWOSTEP method with a heading \ac{RMSE} of $1.42^\circ$. Since the magnetometer measurements in {\bf Sim2} are less \ac{PE} than in {\bf Sim1} the use of the angular velocity measurements allows the MAVBE method to perform better than the magnetometer only TWOSTEP method.

{\bf Sim2} illustrates the main benefit of the proposed MAVBE. The modest attitude motion is sufficient for the MAVBE bias estimates to converge to their true values. However, in {\bf Sim2}, the TWOSTEP method is unable to estimate all of the magnetometer bias terms accurately.

The proposed MAVBE estimator's ability to converge to the proper biases while the instrument experiences modest changes in roll and pitch allows the method to be employed on large full-scale \acp{ROV} and \acp{AUV} which are typically very stable in roll and pitch, and hence, are unable to achieve larger roll and pitch changes required by many common methods like the TWOSTEP method for magnetometer calibration.
\tfflag{R3-2} An extensive investigation of the PE requirement and how to improve
convergence rate would be an interesting future study, but such
studies are beyond the scope of the present paper.

\subsection{Batch vs Continuous Processing and Computational Complexity}
\tfflag{R2-7}
The MAVBE approach can be applied continusouly in real-time or in post-processing.
The TWOSTEP approach is applied in post-processing to a complete data set. 
The computational complexity of the MAVBE approach is
linear in the number of instrument observations and linear in the number of process
model updates, thus its computational complexity is proportional to the temporal
length of a data set.
The computational complexity of the iterative TWOSTEP algorithm, varies both
with the size of the sample data set and the number of algorithm
interations.
Dinale observed that the processing time of ``TWOSTEP on the other hand
seems to dramatically increase the number of iterations it takes to
converge once the sample size increases above 2000 samples''
\citep{dinale2013}.

\section{MAVBE Laboratory Experimental Evaluation}\label{sec:tank}

\subsection{Experimental Test Facility} \label{sec:facility}
Experimental trials were performed with the \ac{JHU} \acf{ROV}, equipped with a MicroStrain 3DM-GX5-25 \citep{microstraingx3datasheet}, in the 7.5 m diameter, 4 m deep fresh water test tank in the  \ac{JHU} \ac{HTF}. The \ac{ROV} is a fully actuated six-\ac{DOF} vehicle with six 1.5 kW DC brushless electric thrusters and employs a suite of sensors commonly employed on deep submergence underwater vehicles.  This includes a high-end \ac{INS}, the iXBlue PHINS III (iXBlue SAS, Cedex, France) \citep{phins.specs,ixsea.phins3.manual}, that is used as a ``ground-truth'' comparison during the experimental trials. The PHINS is a high-end \ac{INS} ($\sim$\$120k) with roll, pitch, heading accuracies of $0.01^\circ$, $0.01^\circ$, $0.05^\circ/\cos$(latitude), respectively \citep{phins.specs}. Figure \ref{fig:tank} shows the \ac{JHU}-\ac{ROV} operating in the test tank.

\begin{figure*}[t!]
	\centering
	\begin{minipage}{0.5\textwidth}\subfigure[{\bf Exp1:} Corrected magnetic field vector]{
			\includegraphics[width=\textwidth]{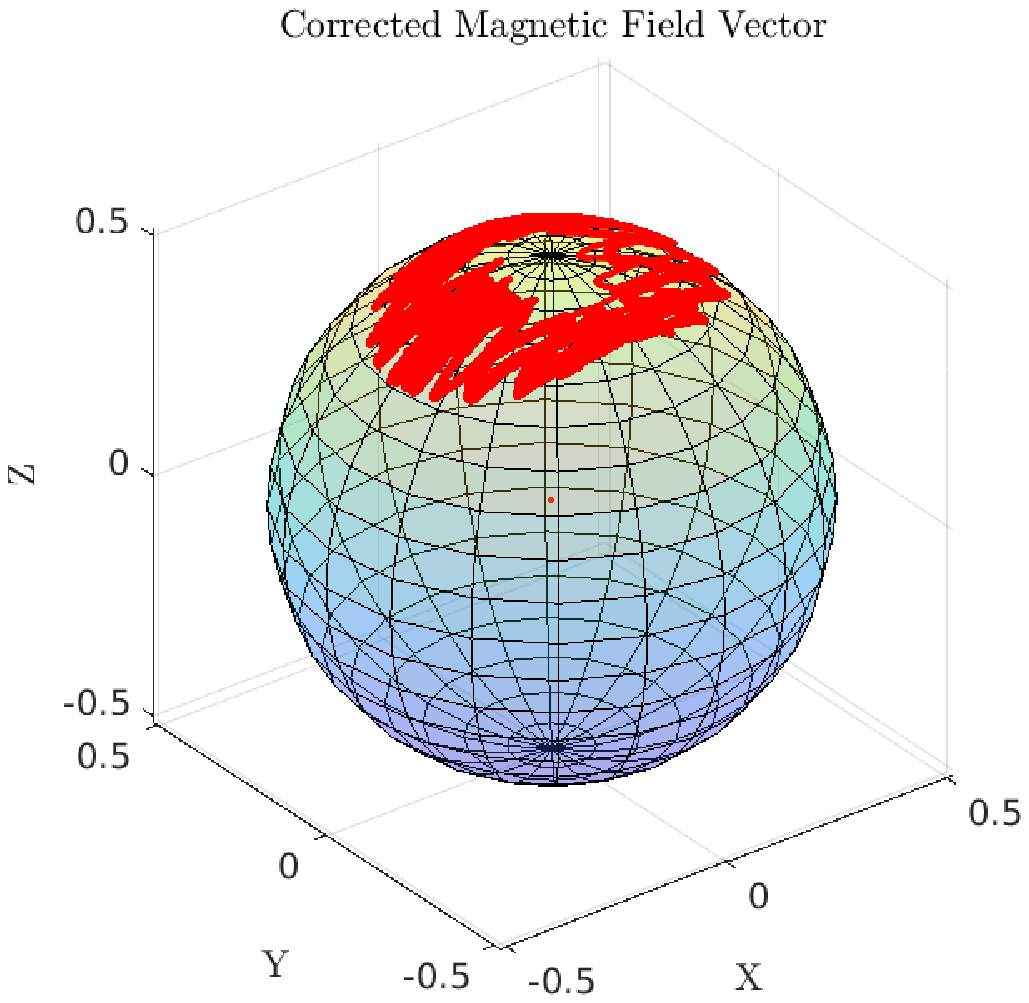}}
	\end{minipage}%
	\hfill
	\begin{minipage}{0.5\textwidth}\subfigure[{\bf Exp2:} Corrected magnetic field vector]{
			\includegraphics[width=\textwidth]{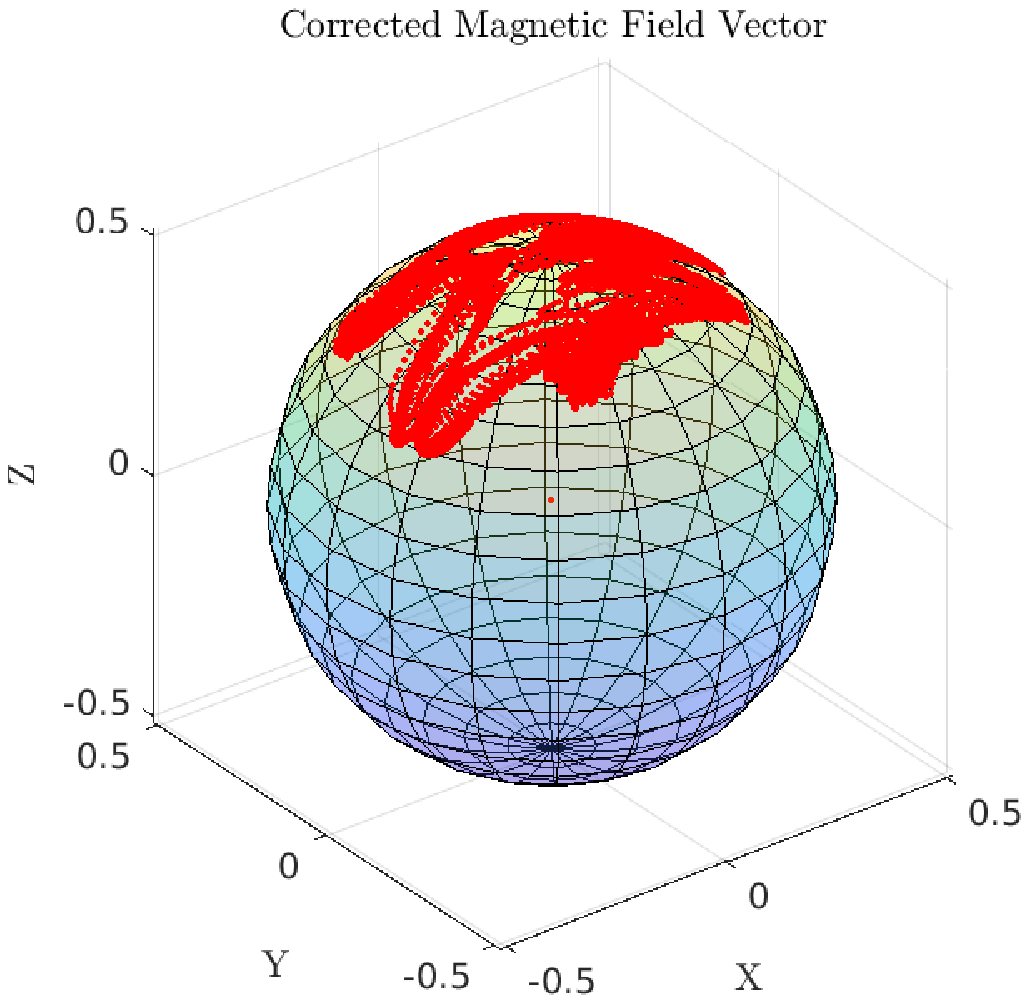}}
	\end{minipage}%
	\caption{{\bf Laboratory Experiments:} The corrected magnetic field vectors for {\bf Exp1} and {\bf Exp2}}.
	\label{fig:exp_mag} 
\end{figure*}

\begin{figure*}[t!]
	\centering
	\includegraphics[width=\textwidth]{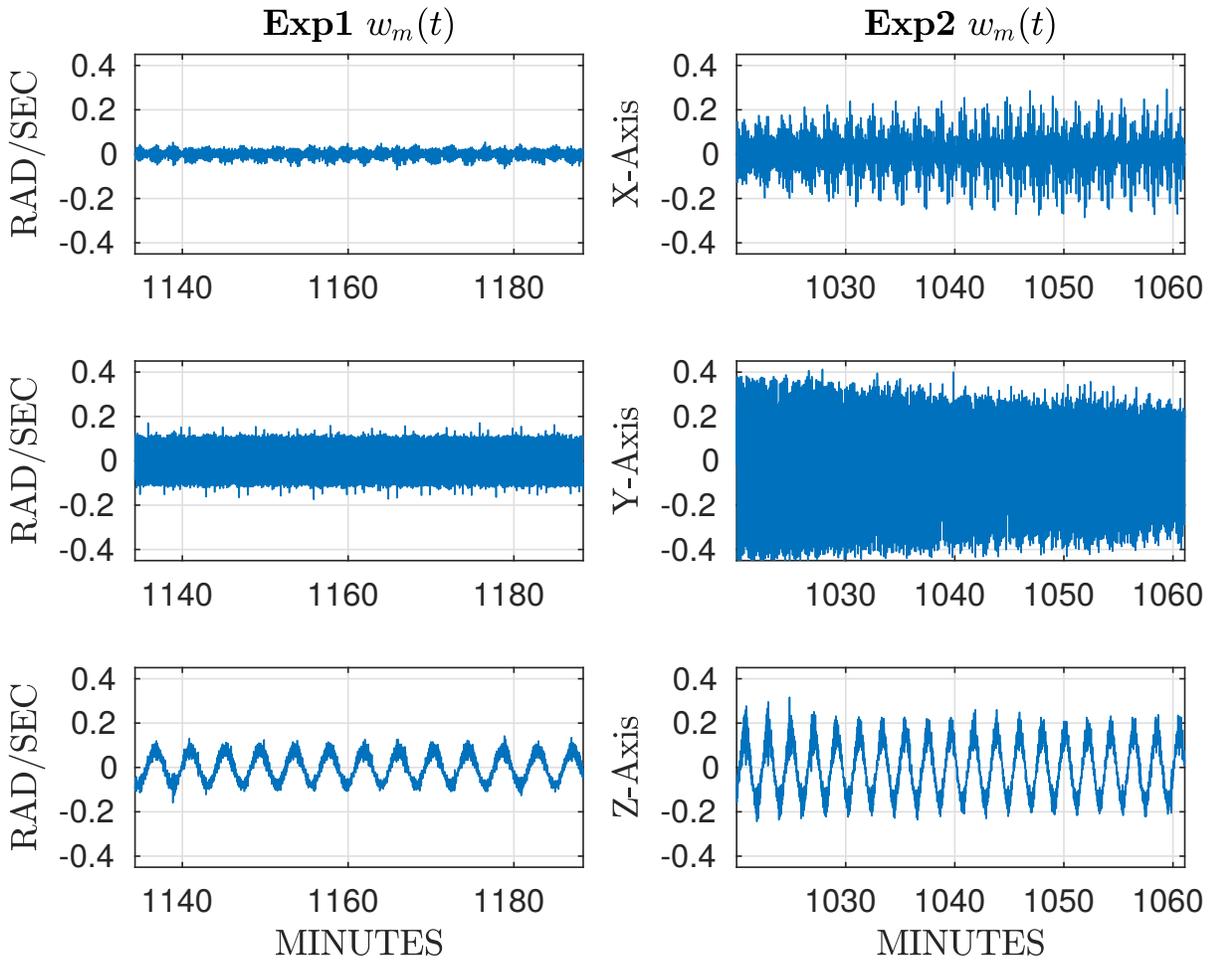}
	\caption{{\bf Laboratory Experiments:} {\bf Exp1} and {\bf Exp2} angular velocity measurements. }
	\label{fig:exp_ang} 
\end{figure*}

\begin{figure*}[t!]
	\centering
	\includegraphics[width=.77\textwidth]{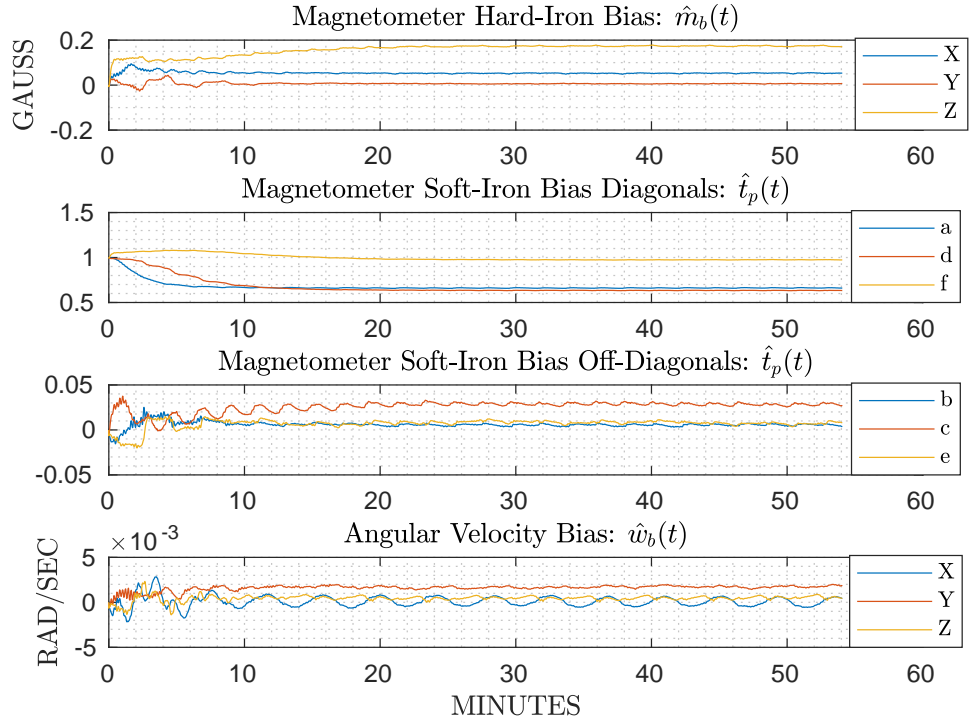}
	\caption{{\bf Experiment 1 (Exp1) Results:} MAVBE magnetometer and angular rate sensor bias estimates converge to constant values. }
	\label{fig:exp1_biases} 
\end{figure*}

\begin{figure*}[t!]
	\centering
	\includegraphics[width=.77\textwidth]{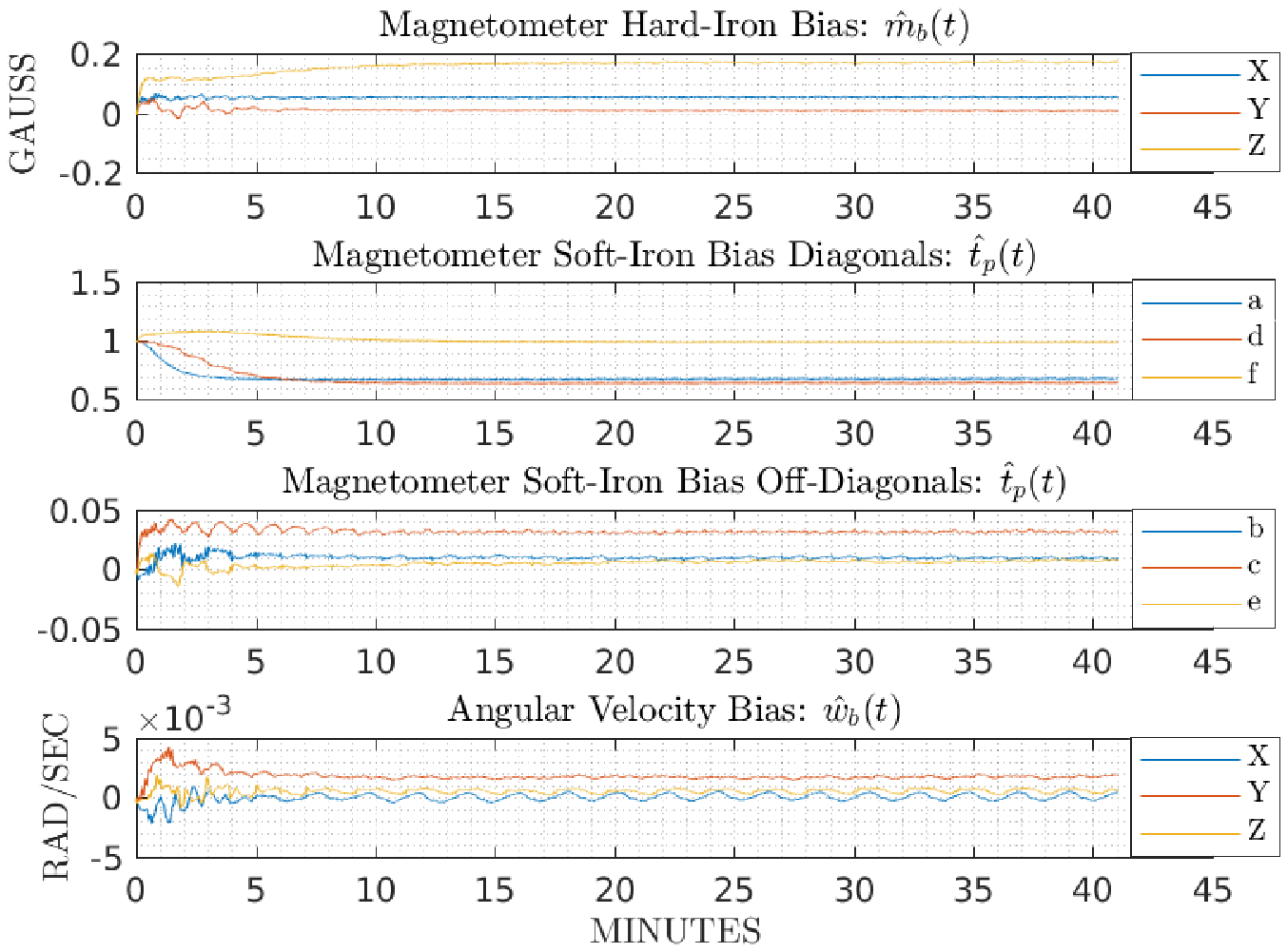}
	\caption{{\bf Experiment 2 (Exp2) Results:} MAVBE Magnetometer and angular rate gyro bias estimates converge to constant values.}
	\label{fig:exp2_biases}
\end{figure*}

\subsection{Experimental Setup}
The \ac{JHU} \ac{ROV} was commanded to execute smooth sinusoidal rotations of roughly $\pm 180^\circ$ in heading and $\pm15^\circ$ in pitch. Two experiments were conducted at different angular velocities. In experiment 1 ({\bf Exp1}), the vehicle was subject to smaller angular velocities than in experiment 2 ({\bf Exp2}). Figure \ref{fig:exp_ang} presents the measured angular velocities from the two laboratory experiments,  and Figure \ref{fig:exp_mag} shows the corrected magnetic field vector during the {\bf Exp1} and {\bf Exp2} experiments.
IMU data was logged at $20$ hz using the MicroStrain 3DM-GX5-25. The initial conditions for the sensor bias estimates are the same as given in Table \ref{table:sim_setup}. The MAVBE was executed at $10$ hz, and the process and measurement covariance matrices are given by (\ref{eq:Q}) and (\ref{eq:R}), respectively.

\subsection{Experimental Results}
The sensor bias estimates for {\bf Exp1} and {\bf Exp2} are presented in Figures \ref{fig:exp1_biases}-\ref{fig:exp2_biases} and the final bias estimates in Table \ref{table:biases_tank}. The results show that the MAVBE sensor bias estimates converge to constant values.
Since the true bias values are unknown, the final bias estimates were used for calibration of the magnetometer and angular rate sensors, enabling heading error to be used as an error metric.  Table \ref{table:rms_tank} reports the MAVBE's and TWOSTEP's calibrated magnetometer respective heading errors. In both laboratory experiments, the heading estimates corresponding to the MAVBE calibrated magnetometer closely tracks the ground truth value with a \ac{RMSE} of roughly $1^\circ$. 
\begin{table*}[t!]
	\centering
	\caption{Estimated bias values for the two laboratory experiments.} 	\footnotesize
	\begin{tabular}{|l | c c c|}
		\hline
		& $m_b$ (G)& $t_p$ & $w_b$ ($^\circ$/s) \\
		\hline
		\bf MAVBE Exp1 & [0.053 0.006 0.171] & [0.662 0.005 0.028 0.634 0.009 0.974] & [0.0006 0.0019 0.0006]\\
		\bf TWOSTEP Exp1 & [0.123 0.002 0.874] & [0.346 0.003 0.017 0.327 0.006 0.395] & N/A \\
		\hline
		\bf MAVBE Exp2 & [0.054 0.008 0.171] & [0.681 0.009 0.031 0.645 0.007 0.990] & [0.0002 0.0018 0.0005]\\
		\bf TWOSTEP Exp2 & [0.106 0.024 0.070] & [0.629 0.010 0.012 0.600 -0.001 1.025] & N/A \\
		\hline
	\end{tabular}
	\label{table:biases_tank}
\end{table*}
\begin{table*}[t!]
	\centering
	\caption{Comparison of heading \acp{RMSE} between different calibration techniques for the two laboratory experiments.}
	\footnotesize
	\begin{tabular}{|c | c c|}
		\hline
		\bf Calibration Method & \bf Exp1 & \bf Exp2 \\
		\hline
		\bf Uncalibrated &  $23.34^\circ$ & $28.00^\circ$ \\
		\bf MAVBE & $0.75^\circ$ & $1.12^\circ$  \\
		\bf TWOSTEP &$2.88^\circ$ & $2.60^\circ$ \\
		\hline
	\end{tabular}
	\label{table:rms_tank}
\end{table*}
However, the TWOSTEP calibrated magnetometer leads to a worse \ac{RMSE} of $>2.5^\circ$. The difference between the \ac{RMSE} corresponding to the MAVBE and TWOSTEP calibration methods demonstrates the advantage of the MAVBE method for providing accurate magnetometer calibration on field vehicles which are unable to achieve the large roll and pitch changes required by many common methods like the TWOSTEP and ellipsoid fitting methods for magnetometer calibration.

In addition, note that in Figures \ref{fig:exp1_biases}-\ref{fig:exp2_biases}, the sensor bias estimates in {\bf Exp2} converged faster than those in {\bf Exp1}. This is due to the fact that the instrument in {\bf Exp2} experienced higher angular velocities than in {\bf Exp1} (seen in Figure \ref{fig:exp_ang}). The increased excitement of the instrument in {\bf Exp2} leads to a faster convergence time for the sensor bias estimates than those in {\bf Exp1}.


\begin{table*}[t!]
	\caption{\ac{JHU} Iver3 Measurement sources, resolutions, and accuracies \citep{microstraingx3datasheet,iver3.manual,os5000,iverdvl}.}
	\begin{center}
		\begin{tabular}{cccc}
			&&&Measurement \\
			State&Source&Update Rate& Resolution \\
			\hline \hline
			Roll, Pitch, Hdg.       & OceanServer OS5000 & \unit{1}{\Hz}& $<0.5^\circ$ Heading RMSE When Level \\
			&Compass Unit&&$1^\circ$ Heading RMSE When $<\pm30^\circ$ Tilt  \\
			\hline
			&Microstrain&&\\
			Linear Acceleration      & 3DM-GX5-25 & \unit{20}{\Hz} &  0.0075 m/s$^2$ Std Dev\\
			\hline
			&Microstrain&&\\
			Angular Velocity           & 3DM-GX5-25 & \unit{20}{\Hz} & 0.00025 rad/s Std Dev\\
			\hline
			Magnetic&Microstrain&&\\
			Field            & 3DM-GX5-25 & \unit{20}{\Hz} & 0.002 gauss Std Dev\\
			\hline
			Translational                   & \unit{600}{\kHz}  & &\\  
			Velocity                & RDI Explorer \acs{DVL}& \unit{4}{\Hz} & 0.01 m/s Std Dev\\
			\hline \hline\end{tabular}
	\end{center}
	\label{table:iver3}
	\vspace*{-5mm}
\end{table*}

\section{MAVBE Field Trial Experimental Evaluation}\label{sec:field}
\subsection{Test Vehicle} \label{sec:facility_iver}
Experimental fields trials were performed with the \ac{JHU}'s Iver3 \ac{AUV} (L3 OceanServer, Fall River, MA, USA) \citep{iver3.manual} in the Chesapeake Bay, MD, USA. The \ac{AUV} is an under-actuated \ac{AUV} equipped with a 600 kHz Phased Array RDI Explorer \ac{DVL} \citep{iverdvl}, a MicroStrain 3DM-GX5-25 \citep{microstraingx3datasheet} \ac{IMU}, and the a OceanServer OS5000 \citep{os5000} magnetic compass. Figure \ref{fig:iver} shows the \ac{JHU} Iver3 \ac{AUV} during the vehicle tests. Table \ref{table:iver3} lists the sensors on board the \ac{JHU} Iver3.

\begin{figure*}[t!]
	\centering
	\begin{minipage}{0.5\textwidth}\subfigure[{\bf Dive1:} Corrected magnetic field vector.]{
			\includegraphics[width=\textwidth]{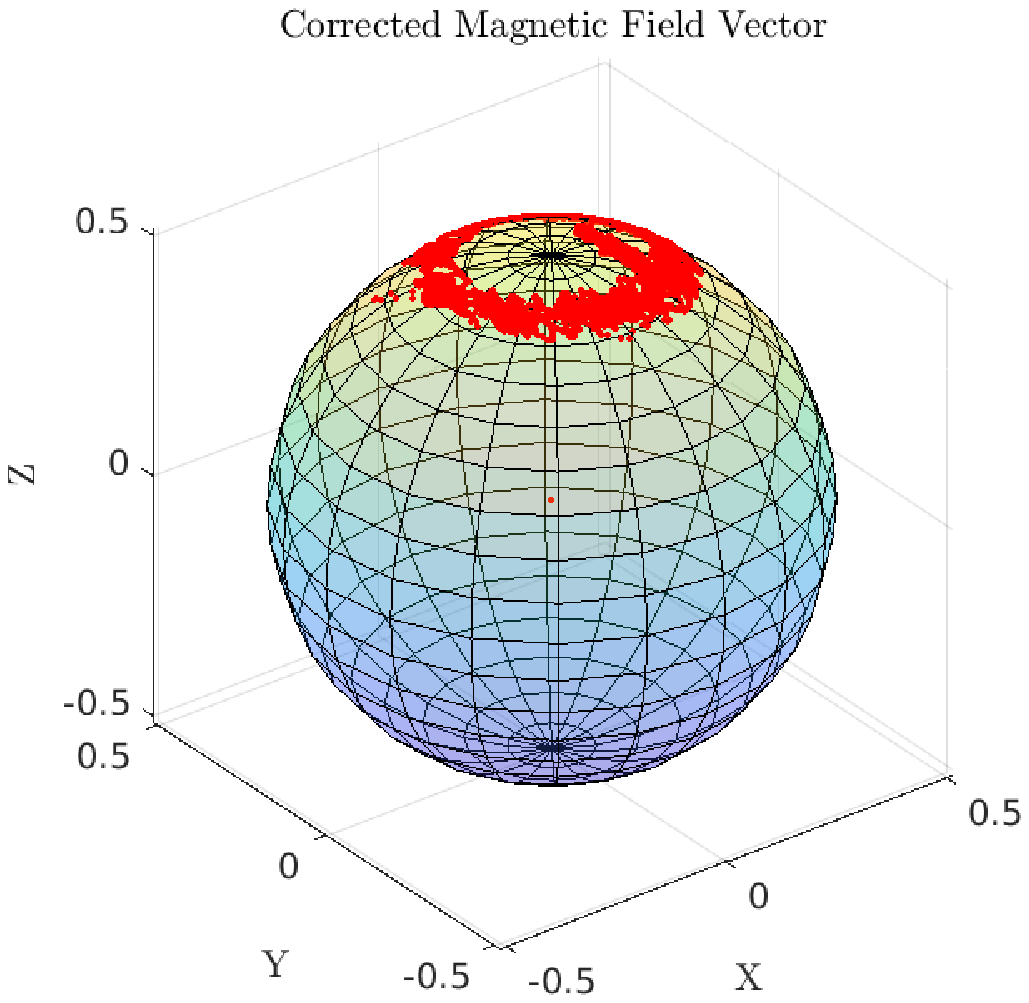}}
	\end{minipage}%
	\hfill
	\begin{minipage}{0.5\textwidth}\subfigure[{\bf Dive2:} Corrected magnetic field vector.]{
			\includegraphics[width=\textwidth]{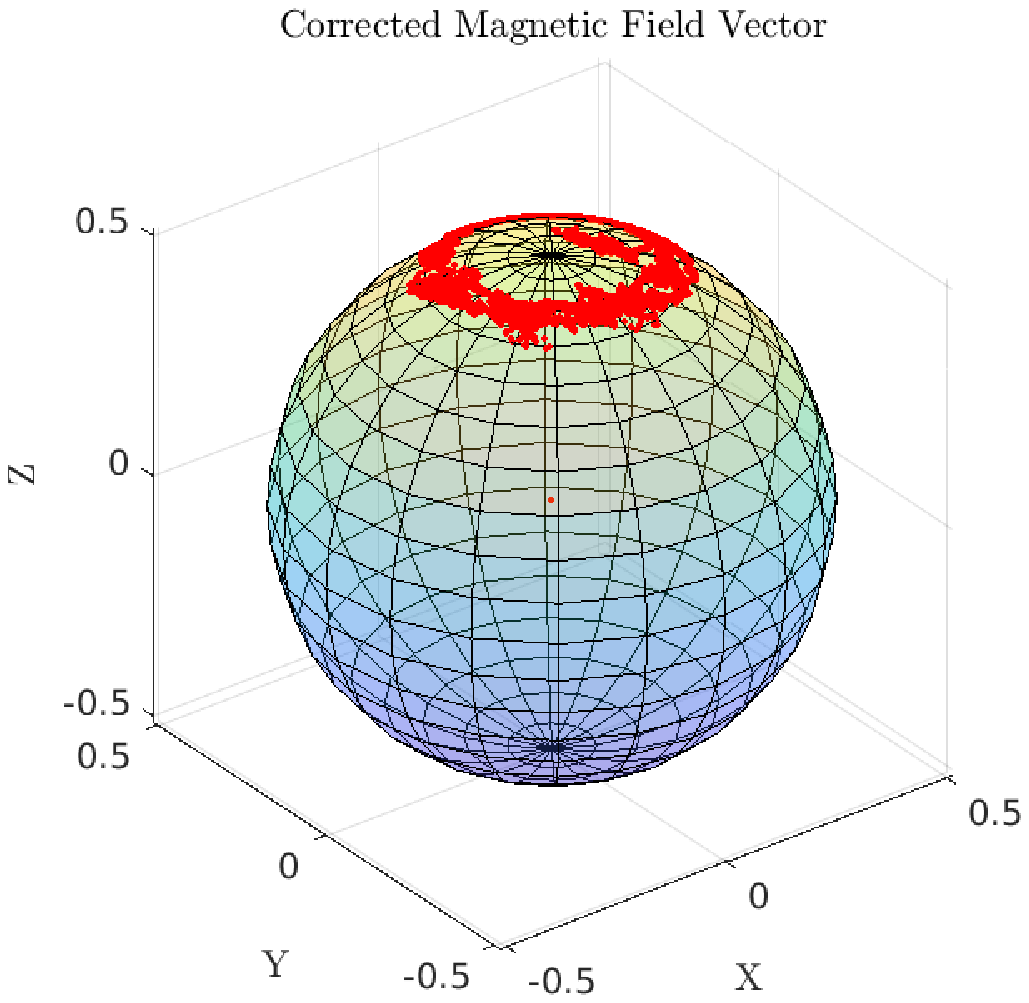}}
	\end{minipage}%
	\caption{{\bf Field Trials:} The corrected magnetic field vectors for {\bf Dive1} and {\bf Dive2}}
	\label{fig:iver_mag}
\end{figure*}

\begin{table*}[t]
	\centering
	\caption{Table of estimated bias values for the two field experiments.} 	\footnotesize
	\begin{tabular}{|l | c c c|}
		\hline
		& $m_b$ (G)& $t_p$ & $w_b$ ($^\circ$/s) \\
		\hline
		\bf MAVBE Dive1 & [-0.084 0.149 0.0146] & [0.986 0.002 -0.001 0.972 -0.002 0.955] & [0.0011 -0.0009 0.0005] \\
		\bf TWOSTEP Dive1 & [-0.085 0.1468 -0.005] & [-0.009 -0.006 -0.067 -0.006 -0.071 -0.990] &  N/A\\
		& + [0.074i 0.078i 1.088i] & + [0.003i 0.003i 0.037i 0.003i 0.039i 0.539i] &  \\
		\hline
		\bf MAVBE Dive2 & [-0.084 0.149 0.020] & [0.970 0.002 -0.002 0.966 -0.003 0.947]  & [0.0012 -0.0008 0.006]\\
		\bf TWOSTEP Dive2 & [0.080 0.022 0.009] & [-0.329 -0.017 0.067 -0.357 0.032 -0.992] & N/A \\
		& + [-0.057i -0.029i 0.560i] & + [0.008i 0.004i -0.077i 0.002i -0.040i 0.754i] & \\
		\hline
	\end{tabular}
	\label{table:biases_iver}
\end{table*}

\subsection{Experimental Setup}
The initial conditions for the sensor bias estimates are given in Table \ref{table:sim_setup}, the same as in the simulations.

The 3DM-GX5-25 \ac{IMU} was sampled at $20$ Hz. The MAVBE was executed at $10$ hz, and the measurement covariance matrix, $R$, was populated with the square of the $\sigma_m$ standard deviations along the diagonal entries such that $R$ is (\ref{eq:R}), and the process covariance matrix, $Q$, is (\ref{eq:Q}).

Before running the two field trials, a compass calibration for the Iver3's OS5000 magnetic compass \citep{os5000} was completed per the instructions by L3 OceanServer \citep{iver3.manual}. Note that the \ac{JHU} Iver3 \ac{AUV} has its own L3 OceanServer proprietary magnetometer calibration method for the OS5000 magnetic compass based on heading sweeps and a look up table. This process is a two step process involving heading sweeps on a stand and an in-water compass calibration mission. During the field trials, the proposed MAVBE method is compared not only to the TWOSTEP method (as done in the simulation and laboratory experiments), but also to OceanServer's OS5000 commercial magnetic compass.

\begin{figure*}
	\centering
	\includegraphics[width=.77\textwidth]{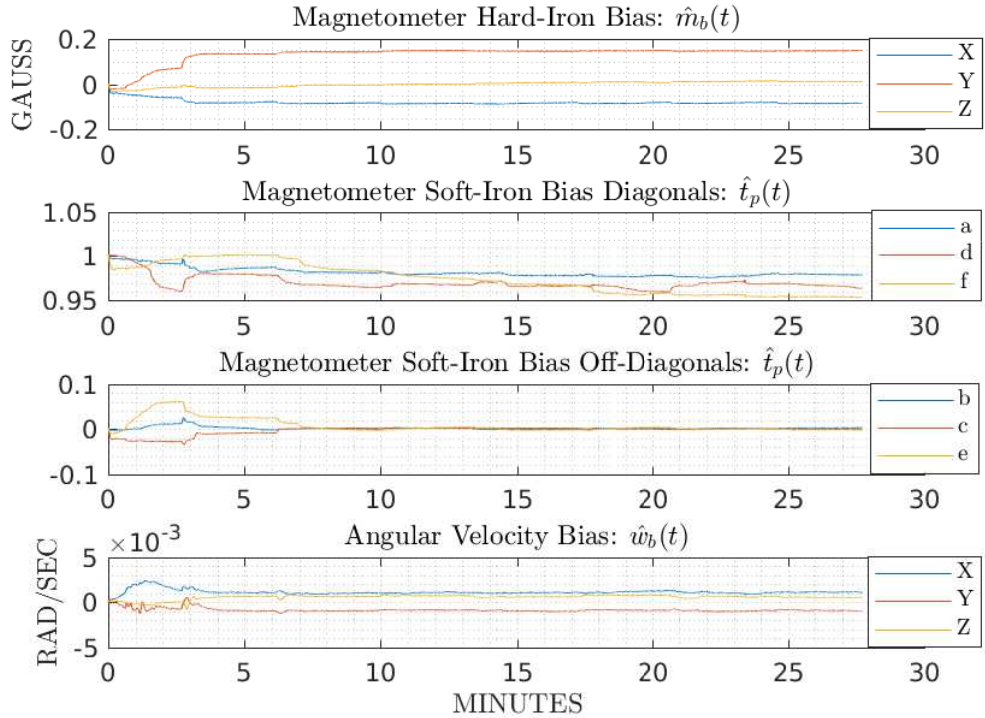}
	\caption{{\bf Dive 1 (Dive1) Results:} MAVBE magnetometer and angular rate sensor bias estimates converge to constant values.}
	\label{fig:dive1_bias}
\end{figure*}

\begin{figure*}
	\centering
	\includegraphics[width=.77\textwidth]{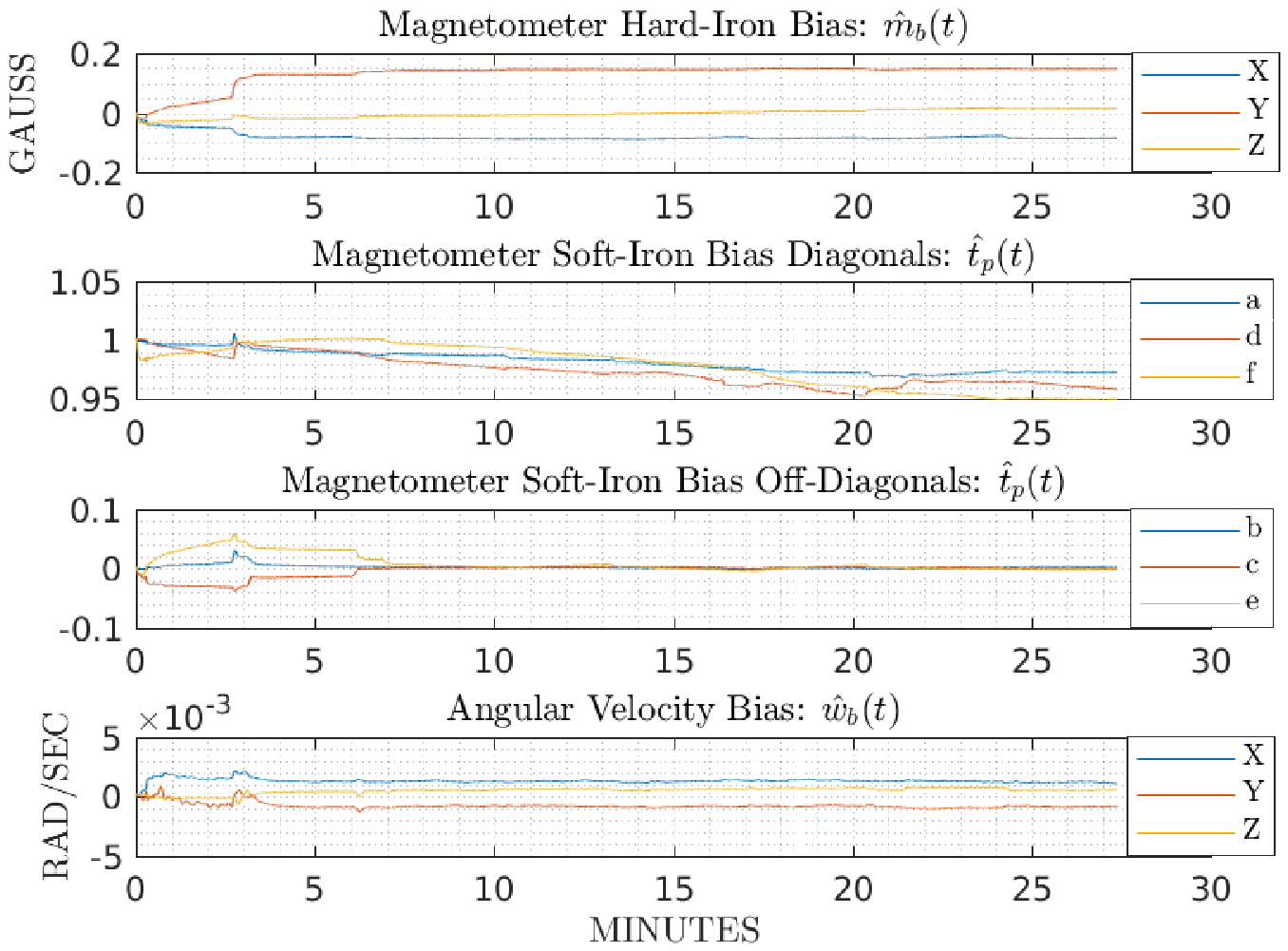}
	\caption{{\bf Dive 2 (Dive2) Results:} MAVBE magnetometer and angular rate sensor bias estimates converge to constant values.}
	\label{fig:dive2_bias}
\end{figure*}

\begin{figure*}
	\centering
	\includegraphics[width=.72\textwidth]{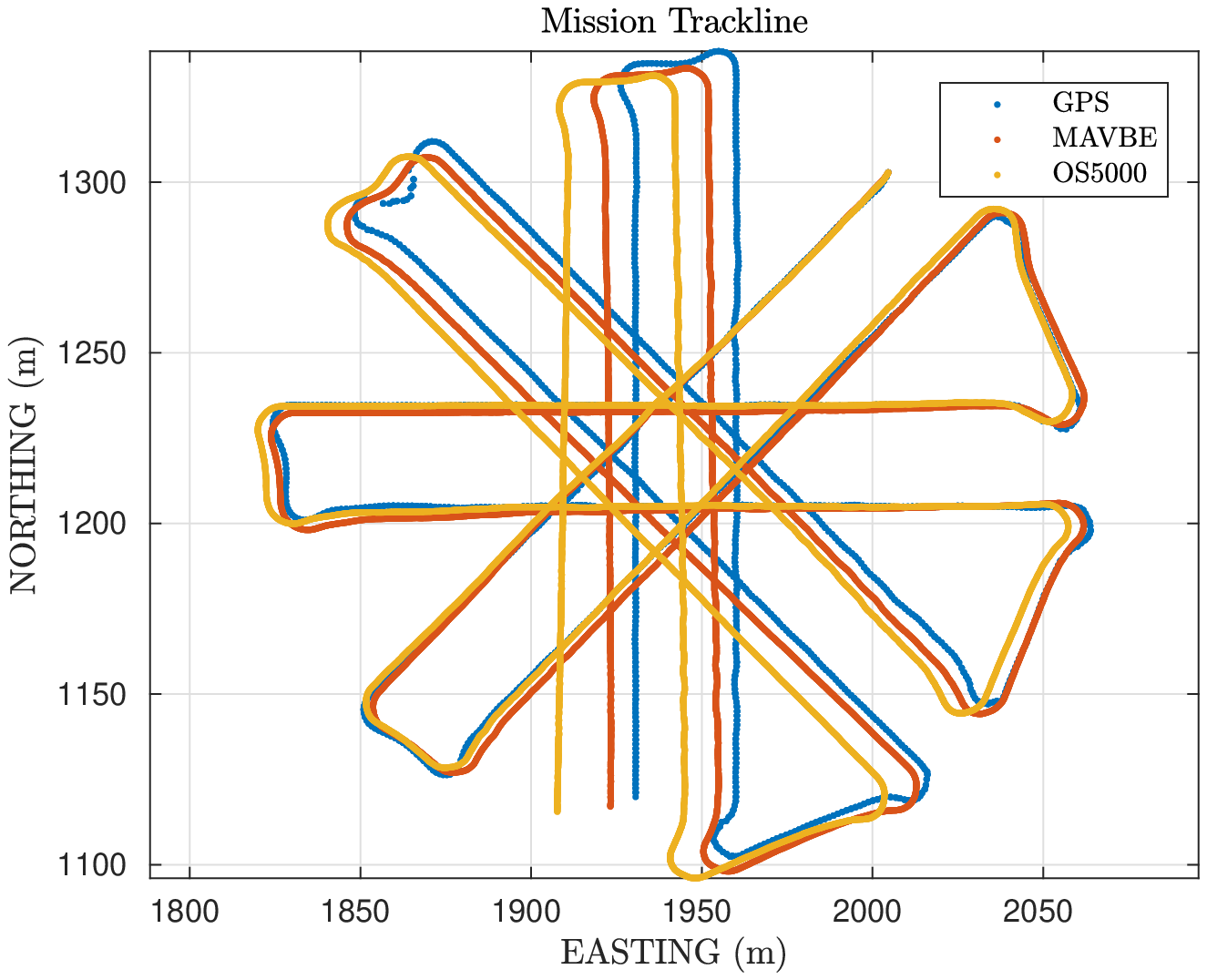}
	\caption{{\bf Dive 1 (Dive1) Navigation Tracks}: Comparison of the Doppler dead reckoning navigation between the tracks from MAVBE compass calibration, the OS5000 calibrated compass, and the \ac{GPS} ground truth track.}
	\label{fig:dive1_track}
\end{figure*}

\begin{figure*}
	\centering
	\includegraphics[width=.72\textwidth]{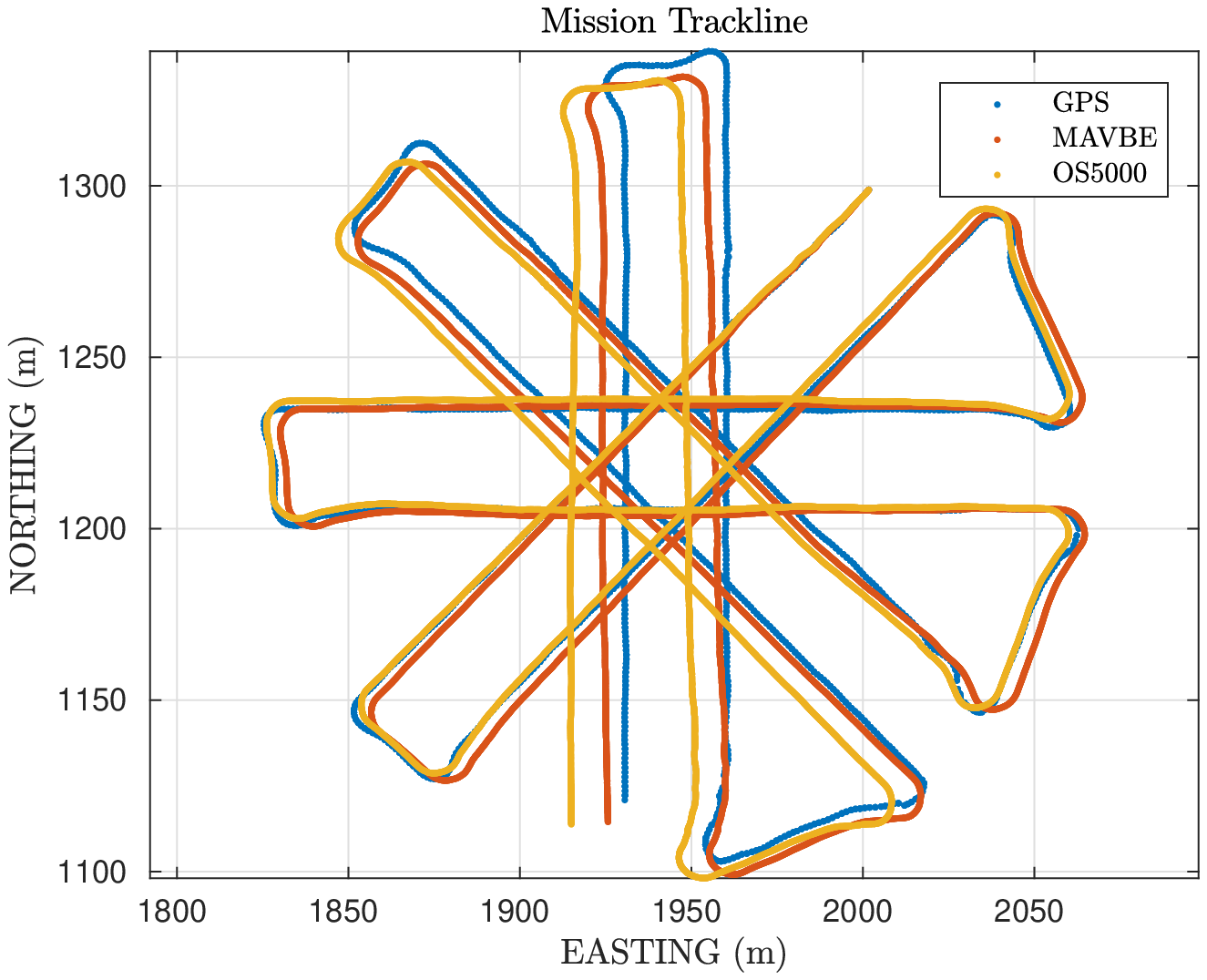}
	\caption{{\bf Dive 2 (Dive2) Navigation Tracks}: Comparison of the Doppler dead reckoning navigation between the tracks from MAVBE compass calibration, the OS5000 calibrated compass, and the \ac{GPS} ground truth track.}
	\label{fig:dive2_track}
\end{figure*}

\begin{figure*}
	\centering
	\includegraphics[width=.72\textwidth]{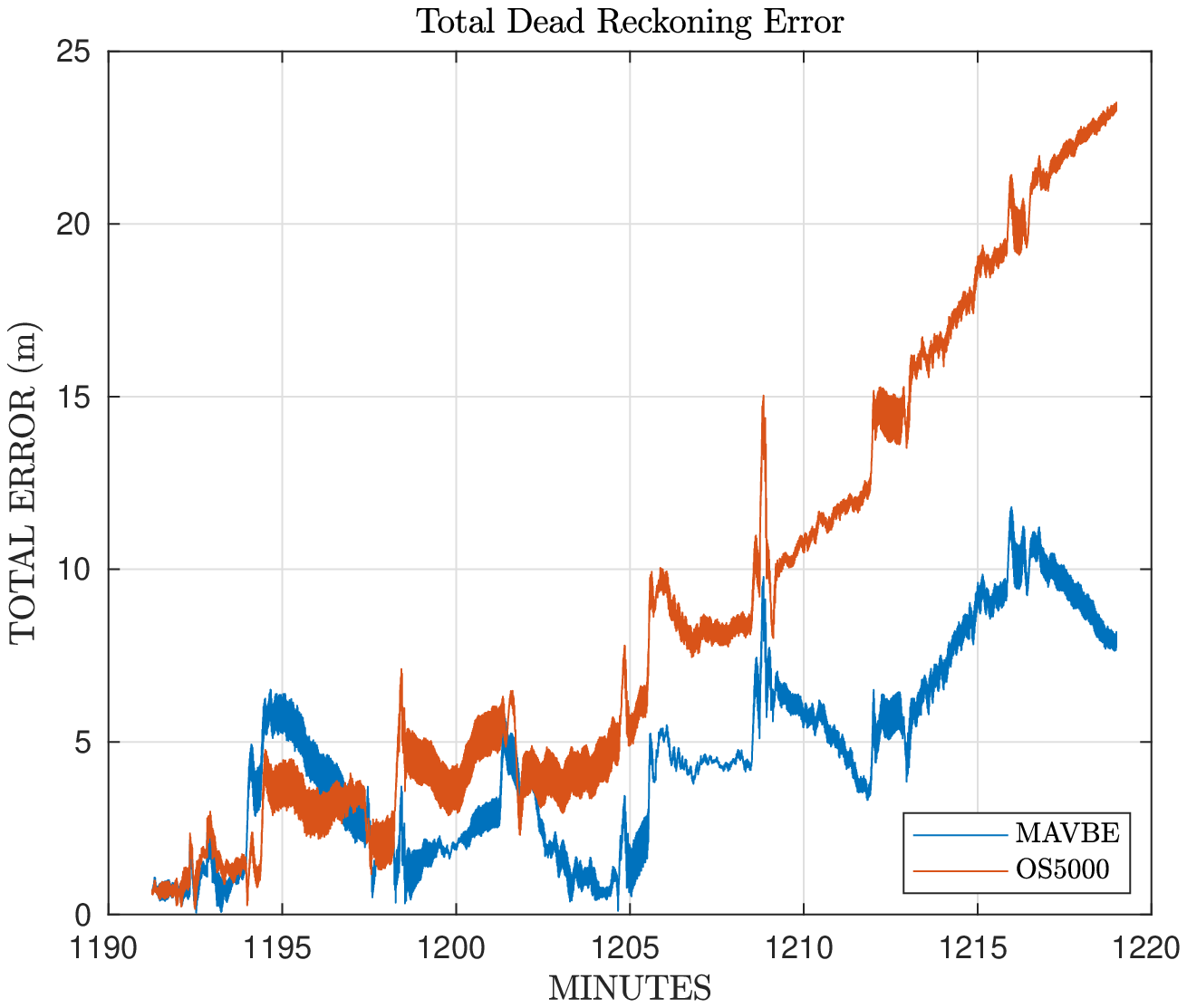}
	\caption{{\bf Dive 1 (Dive1) Navigation Error}: Comparison of the Doppler dead reckoning navigation error of the tracks from MAVBE compass calibration and the OS5000 calibrated compass.}
	\label{fig:dive1_err}
\end{figure*}

\begin{figure*}
	\centering
	\includegraphics[width=.72\textwidth]{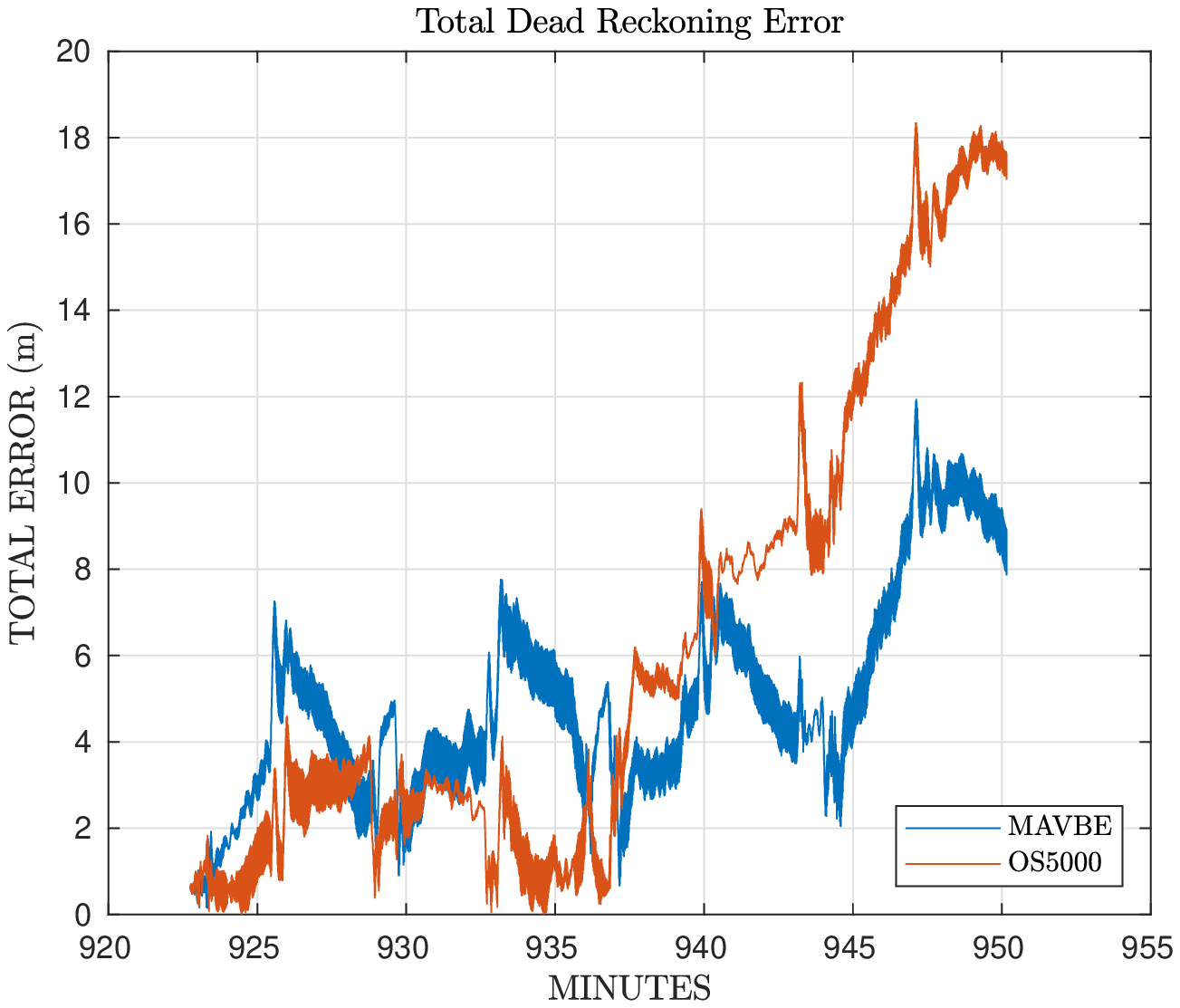}
	\caption{{\bf Dive 2 (Dive2) Navigation Error}: Comparison of the Doppler dead reckoning navigation error of the tracks from MAVBE compass calibration and the OS5000 calibrated compass.}
	\label{fig:dive2_err}
\end{figure*}

Two field trials were conducted in Round Bay on the Chesapeake Bay, MD, USA. The \ac{JHU} Iver3 \ac{AUV} conducted two surface missions following cardinal and intercardinal heading directions. The trials were designed so that the \ac{JHU} Iver3 \ac{AUV} would follow track-lines along cardinal and intercardinal heading directions in order to ensure that the magnetometer is fully calibrated for all heading directions. The trials were conducted on the water surface to allow for \ac{GPS} to be used as a navigation ground truth and in shallow water to allow for \ac{DVL} bottom-lock velocity measurements. Two trials were conducted $\sim 4.5$ hours apart (one in the morning and one in the afternoon) in order to confirm our assumption that the sensor biases were in fact slowly time varying and did not change drastically between experiments. Since the \ac{JHU} Iver3 does not have a high-end \ac{INS} (like used on the \ac{JHU} \ac{ROV} in the laboratory experiments) to compare heading estimates, navigation error is used as a proxy for the magnetometer calibration during the field trials. The recalculated dead reckoning navigation for each magnetometer calibration method is compared to the \ac{GPS} track of the vehicle.



\subsection{Experimental Results}

The MAVBE's sensor bias estimates for {\bf Dive1} and {\bf Dive2} are
presented in Figures \ref{fig:dive1_bias} - \ref{fig:dive2_bias} and
Table \ref{table:biases_iver}. The results show that the MAVBE's
sensor bias 
estimates \tfflag{R1-2} evolve from nominal initial values to converged
values within about 5-10 minutes.
\tfflag{R3-2} This also demonstrates that  the motion of an actual AUV on a survey mission
trajectory results in the required PE needed for convergence of the
MAVBE parameter estimates.
The final bias
estimates were used for calibration of the magnetometer and angular
rate sensor. Table \ref{table:biases_iver} reports the final estimated
sensor biases from {\bf Dive1} and {\bf Dive2} used for
calibration. Figure \ref{fig:iver_mag} shows the corrected magnetic
field vectors from the two field experiments. From Figure
\ref{fig:iver_mag}, it is evident that the Iver3 \ac{AUV} experienced
very little excitement in roll and pitch, and hence there is not great
coverage of the sphere.  As discussed earlier, common methods for
magnetometer calibration like the TWOSTEP and ellipsoid fitting
methods require sufficient excitement of the magnetometer for their
convergence to the correct sensor bias estimates. During the two field
experiments however, the \ac{AUV} was stable in roll and pitch, and
the TWOSTEP method was unable to accurately estimate the sensor bias.
Table \ref{table:biases_iver} reports the TWOSTEP estimated biases,
which are clearly incorrect as they are found to be imaginary.

As reported in Table \ref{table:biases_iver}, the TWOSTEP algorithm failed to produce realistic bias estimates for either {\bf Dive1} or {\bf Dive2}. For these dives, the TWOSTEP method produced (physically meaningless) imaginary values for the magnetometer hard-iron and soft-iron estimates.  Hence, the proposed MAVBE calibration is compared to the Iver3 \ac{AUV}'s calibrated OS5000 compass which relies on L3 OceanServers proprietary magnetometer calibration method based on heading sweeps and a look up table. In addition, since the Iver3 \ac{AUV} does not have a high end \ac{INS} to provide heading as a comparison to the calibrated magnetometer heading, Doppler navigation error is used as an error metric. 

Figures \ref{fig:dive1_track}-\ref{fig:dive2_track} show the MAVBE and
OS5000 calibrated magnetic compass navigation tracks in comparison
with the \ac{GPS} ground truth track, and Figures
\ref{fig:dive1_err}-\ref{fig:dive2_err} show the the respective two
norm error of the navigation tracks. From Figures
\ref{fig:dive1_err}-\ref{fig:dive2_err}, we see that the MAVBE
calibrated compass leads to more accurate Doppler dead reckoning
navigation than the industry standard provided by L3 OceanServer's
calibrated OS5000 magnetic compass. The navigation track error for the
{\bf Dive1} and {\bf Dive2} field experiments demonstrate that the
MAVBE calibrated MicroStrain 3DM-GX5-25 provides improved performance
over the calibrated OS5000 compass. In addition, the field experiments
demonstrate the ability of the MAVBE method to properly calibrate
magnetometers on robotic vehicles which are stable in roll and pitch and,
in consequence, experience only modest excurions in roll and pitch.
This is in contrast to the TWOSTEP method which is unable to estimate
the magnetometer hard-iron and soft-iron sensor bias during the field
trials.

\section{Conclusion}\label{sec:conc}
This paper reports a novel method for on-line, real-time estimation of hard-iron and soft-iron magnetometer biases and angular rate sensor biases in \acfp{IMU} for use in \acp{AHRS}. \acp{AHRS} commonly use bias-compensated magnetometer measurements to estimate heading. By utilizing angular rate sensor measurements, smaller angular rotations of the instrument (in comparison to previously reported methods for magnetometer calibration) are required for accurate compensation of magnetometer and angular velocity sensor biases. Since the proposed estimator works with smaller changes in roll and pitch than previously reported methods, it can be implemented on real full-scale \acp{ROV} to provide online estimates of magnetometer sensor biases to allow real-time bias-compensation for these sensors.

Oceanographic \acfp{UV} and surface vehicles, which are commonly passively stable in roll and pitch, are unable to achieve the roll and pitch required for the common magnetometer calibration methods like the TWOSTEP and ellipsoid fitting methods.  The simulations, laboratory experiments, and field trials show that the proposed MAVBE magnetometer calibration method provides improved performance over common methods like the TWOSTEP method and the OceanServer Iver3 \ac{AUV} commercial solution.  The TWOSTEP method was unable to estimate the magnetometer sensor biases during the field experiments due to the limited motion in roll and pitch.  The ability of the  proposed MAVBE method to accurately estimate magnetometer biases when there is limited excitation of the magnetometer signal illustrates the advantage of the proposed calibration method over common calibration methods like the TWOSTEP and ellipsoid fitting methods which fail when there is low coverage of the magnetometer on the sphere. 

As demonstrated in the field experiments, the proposed method leads to improved position estimation of the Iver3 \ac{AUV} over the TWOSTEP calibrated magnetometer and the calibrated OceanServer OS5000 magnetic compass.

In future studies, the authors hope to improve the convergence time of the estimator by developing coarse and fine alignment protocols. Different values for the process covariance matrix, $Q$, could be chosen during the coarse and fine alignment to enable both a fast convergence of the sensor biases and a smooth final steady state.

\section*{ACKNOWLEDGMENT}
We gratefully acknowledge the support of the National Science
Foundation under NSF awards OCE-1435818 and IIS-1909182. In addition, the authors acknowledge the masters thesis by Dinale \citep{dinale2013} which provides a clear derivation and well documented reference source code for the TWOSTEP method originally reported by Alonso and Shuster \citep{alonso2002twostep}.



\bibliographystyle{apalike}
\bibliography{abhi_2}

\end{document}